\documentclass[runningheads,a4paper]{llncs}

\usepackage{amssymb}
\usepackage{graphicx}

\usepackage{url}
\urldef{\mailsa}\path|{shgo, veroneze, vonzuben}@dca.fee.unicamp.br|    
\newcommand{\keywords}[1]{\par\addvspace\baselineskip
\noindent\keywordname\enspace\ignorespaces#1}

\usepackage{paralist}
\usepackage{amsmath}
\usepackage{subfig}
\usepackage{float}
\floatstyle{plaintop}
\restylefloat{table}

\begin{document}

\mainmatter  

\title{On bicluster aggregation and its benefits for enumerative solutions}

\titlerunning{On bicluster aggregation and its benefits for enumerative solutions}

%
%
\author{Saullo Oliveira%
\thanks{}%
\and Rosana Veroneze \and Fernando J. Von Zuben}
\authorrunning{On bicluster aggregation and its benefits for enumerative solutions}

\institute{School of Electrical and Computer Engineering,\\
Unicamp, Campinas, S\~{a}o Paulo, Brazil\\
\mailsa \\
\url{http://www.fee.unicamp.br}}

%
%

\toctitle{Lecture Notes in Computer Science}
\tocauthor{Authors' Instructions}
\maketitle

\begin{abstract}
Biclustering involves the simultaneous clustering of objects and their attributes, thus defining
local two-way clustering models.
Recently, efficient algorithms were conceived to enumerate all biclusters in real-valued datasets.
In this case, the solution composes a complete set of maximal and non-redundant biclusters.
However, the ability to enumerate biclusters revealed a challenging scenario: in noisy datasets, each true bicluster may become highly fragmented and with a high degree of overlapping.
It prevents a direct analysis of the obtained results.
Aiming at reverting the fragmentation, we propose here two approaches for properly aggregating the whole set of enumerated biclusters: one based on single linkage and the other directly exploring the rate of overlapping.
Both proposals were compared with each other and with the actual state-of-the-art in several experiments, and they not only significantly reduced the number of biclusters but also consistently increased the quality of the solution.
.
\end{abstract}

\section{Introduction}
Biclustering techniques aim to simultaneously cluster objects and attributes of a dataset.
Each bicluster is represented as a tuple containing a subset of the rows, and a subset of the columns, as long as they exhibit some kind of coherence pattern.
There are several kinds of coherence which can be found in a bicluster, and they directly interfere on the mechanism of bicluster identification.
As finding all biclusters in a dataset is an NP-hard problem, several heuristics were proposed, such as CC \cite{Cheng2000} and FLOC \cite{Yang2003}.
Such heuristics may miss important biclusters, and may also return non-maximal biclusters (biclusters that can be further augmented).

In the case of binary datasets, there are a plenty of algorithms for enumerating all maximal biclusters. Some examples are Makino \& Uno \cite{Makino2014}, LCM \cite{Uno2004} and In-Close2 \cite{Andrews2009}.
The enumeration of all maximal biclusters in an integer or real-valued dataset is a much more challenging scenario, but we already have some proposals, such as RIn-Close \cite{Veroneze2014} and RAP \cite{Pandey2009}.

The drawback of enumerative algorithms, particularly in the context of noisy datasets, is the existence of a large number of biclusters, due to fragmentation of a much smaller number of true biclusters.
This is exemplified in one of our experiments, where we take artificial datasets, gradually increment the variance of a Gaussian noise, and get the enumerative result.
As shown in Fig. \ref{fig:qtds}, with enough noise, the enumerative results exhibit an strong increase on the quantity of biclusters.
This fragmentation leads to a challenging scenario for the analysis of the results, which can become impractical even in small datasets.
In fact, the noise is responsible for fragmenting each true bicluster into many with high overlapping, so that the aggregation of these biclusters is recommended \cite{Liu04} \cite{Zhao2005}.

We propose a way of aggregating biclusters from a biclustering result that shows a high overlapping among its components, as it is the case when enumerating biclusters in noisy datasets.
For this reason, in this paper we will focus on enumerative results, but our proposal can be applied to the result of any algorithm that returns biclusters with high overlapping among them.
The formulation is based on the fact that the high overlapping among biclusters may indicate that they are fragments of a true bicluster that should be reconstructed.
We propose two different techniques to perform the aggregation, followed by a step that removes elements that should not be part of a bicluster.
We performed experiments with three artificial datasets posing different challenges, and two real datasets from distinct backgrounds.
We compared our proposals with a bicluster ensemble algorithm, and the merging/deleting steps of MicroCluster \cite{Zhao2005}.
The experimental results show that the aggregation not only severely reduces the quantity of biclusters, but also tends to increase the quality of the solution.

The paper is organized as follows. In Section \ref{related}, we give the main definitions and discuss the related works in the literature.
Section \ref{proposals} outlines our proposals.
The metrics used to evaluate our proposals will be presented in Section \ref{metrics}.
In Section \ref{experiments}, we present the experimental procedure and the obtained results of the experiments.
Concluding remarks and future work are outlined in Section \ref{conclusions}.

\section{Definitions and Related Work}
\label{related}
Consider a dataset $\textbf{A} \in \mathbb{R}^{n \times m}$, with rows $X = \{x_1, x_2, \dots, x_n\}$ and columns $Y = \{y_1, y_2, \dots, y_m\}$.
We define a bicluster $B = (B^r, B^c)$, where $B^r \subseteq X$ and $B^c \subseteq Y$, such that the elements in the bicluster show a coherence pattern.
A bicluster solution is a set of biclusters represented by $\bar{B} = \{B_i\}_{i=1}^q$, containing $q$ biclusters.
A bicluster is maximal if and only if we can not include any other object / attribute without violating the coherence threshold.
If a solution contains non-maximal biclusters, the result is redundant because there will be biclusters which are part of larger ones.

Madeira \& Oliveira \cite{Madeira2004} categorized the types of biclusters according to their similarity patterns.
They also categorized the biclusters structure in a dataset based on their disposition and level of overlapping.
We highlight that biclusters with constant values, constant values on rows, or constant values on columns are special cases of biclusters with coherent values, and we will focus our attention on the latter, due to its generality.
For a comprehensive survey of biclustering algorithms, the reader may refer to \cite{Madeira2004} and \cite{Tanay2005}.

The overlapping between two biclusters $B$ and $C$ is an important concept in this work, and is defined as:
\begin{equation}
ov(B, C) = \frac{|B^r \cap C^r \times B^c \cap C^c|}{min(|B^r \times B^c|, |C^r \times C^c|)}.
\end{equation}

Now we shall proceed to the aggregation proposals in the literature.
It is important to highlight that, when aggregating two maximal biclusters, the coherence threshold will be violated.
Otherwise, the biclusters would not be maximal.

\subsection{MicroCluster Aggregation}
\label{cap:ens:aggregation:mc}
MicroCluster \cite{Zhao2005} is an enumerative proposal that has two additional steps after the enumeration.
These steps have the task of deleting or merging biclusters which are not covering an area much different from other biclusters.
The first is the deleting step.
If we find a bicluster such that the ratio of its area that is not covered by any other bicluster, by its total area, is less than a threshold $\eta$, it can be removed.
The second step is the merging one.
Let us consider two biclusters and generate a third one with the union of rows and columns of the previous two.
If the ratio of the area of the third bicluster that is not covered by any of the previous two, by its total area, is less than a threshold $\gamma$, we can aggregate the two biclusters into this third one.
In this method of aggregation, non-maximal biclusters will be removed in the deleting step, thus not interfering in the final result.
For more details, please refer to Zhao \& Zaki \cite{Zhao2005}.

\subsection{Aggregation Using Triclustering}
Triclustering was proposed by Haczar \& Nadif \cite{Hanczar2012} as a biclustering ensemble algorithm.
First, they transform each bicluster into a binary matrix.
After that, they propose a triclustering algorithm to find the $k$ most relevant biclusters.
As they were able to improve the biological relevance of biclustering for microarray data \cite{Hanczar2011b}, we will use this method as a contender in this paper.
One major point in ensemble is that we want to combine the results reinforcing the biclusters that seem to be important for several components, and discarding the ones that may come from noise.
Due to the way the triclustering algorithm handles the optimization step, non-maximal biclusters can interfere in the final results.

Bicluster aggregation is slightly different from bicluster ensemble.
While on ensemble tasks we discard biclusters that seem unimportant and combine the ones that contribute the most for the solution, in bicluster aggregation we never discard any bicluster.
Given this characteristic, the bicluster ensemble solution is expected to show a high \textit{Precision} with an impacted \textit{Recall} (see Section \ref{metrics}), as it eliminates biclusters.

\subsection{Other Aggregation Methods}
Gao \& Akoglu \cite{GAO2014} used the principle of Minimum Description Length to propose CoClusLSH, an algorithm that returns a hierarchical set of biclusters.
The hierarchical part can be seen as an aggregation step.
This step is done based on the LSH technique as a hash function.
Candidates hashed to the same bucket are then aggregated until no merging improves the final solution.
Their work is focused in finding biclusters in a checkerboard structure, that does not allow overlapping, thus being not suitable for the kind of problem we are dealing with.

Liu \textit{et al}. \cite{Liu04} proposed OPC-Tree, a deterministic algorithm to mine Order Preserving Clusters (OP-Clusters), a general case of Order Preserving Sub Matrices (OPSM) type of biclusters.
They also have an additional step for creating a hierarchical aggregation of the OP-Clusters.
The Kendall coefficient is used to determine which clusters should be merged and in which order the objects should participate in the resultant OP-Cluster.
The highest the Rank Correlation using the Kendall coefficient, the highest the similarity between two OP-Clusters.
The merging is allowed according to a threshold that is reduced in a level-wise way.
OPC-Tree considers the order of the rows in the bicluster.
In this work, we are dealing with biclusters of coherent values.
In this case, a perfect coherent values bicluster keeps the order of its rows and the hierarchical step of OPC-Tree would be able to be used in this case as well.
But we are considering noisy datasets, in which this assumption probably will not hold, thus the hierarchical step of OPC-Tree is not suitable for the problem we are dealing with.

\section{New Proposals for Aggregation}
\label{proposals}

\subsection{Aggregation with Single Linkage}
\label{aggregation:sl}
Our first proposal receives as input a biclustering solution $\bar{B}$, from enumeration or from a result presenting high overlapping among its components.
With this solution, we transform each bicluster into a binary vector representation as follows:
Given the dimensions of the dataset $\textbf{A} \in \mathbb{R}^{n \times m}$, each bicluster will be a binary vector $\textbf{x}$ of length ${n + m}$.
For a bicluster $B$ transformed into the binary vector $\textbf{x}$, the first $n$ positions represent the rows of the dataset $\textbf{A}$ and if the bicluster contains the $i$th row, $\textbf{x}_i = 1$, otherwise $\textbf{x}_i = 0$.
The last $m$ positions represent the columns of the dataset $\textbf{A}$ and if the bicluster contains the $i$th column, $\textbf{x}_{n+i} = 1$, otherwise $\textbf{x}_{n+i} = 0$.
After this transformation, we use the Hamming distance to apply the single linkage clustering on the existing biclusters.
Notice that the Hamming distance on this transformation will just count how many rows and columns are different among the two biclusters.
In this case, a non-maximal bicluster may be distant from the bicluster that covers its maximal area, thus impacting the quality of the results of this method of aggregation.
In this case, it is necessary that this proposal receives a biclustering solution $\bar{B}$ containing only maximal biclusters.

After choosing a cut on the dendrogram, we aggregate all biclusters that belong to a junction using the function \textit{aggreg}, defined as:
\begin{equation}
aggreg(B, C) = (B^r \cup C^r, B^c \cup C^c),
\label{eq:aggreg}
\end{equation}
that is simply the union of rows / columns of the biclusters.
It is important to note that the \textit{aggreg} function is associative, since it is based on the union operation.
Moreover, we want to highlight that the direct union of rows / columns may include elements that should not be part of a bicluster.
In Section \ref{aggregation:outlier} we will present a way to remove rows / columns that may be interpreted as outliers.

\subsection{Aggregation by Overlapping}
\label{aggregation:by_ov}
It seems intuitive to aggregate the biclusters with an overlapping rate above a defined threshold.
This proposal is based on the aggregation by pairs: while having two biclusters with an overlapping rate higher than a pre-determined threshold $th$, we remove them from the set of biclusters, and include the result of the function \textit{aggreg}, defined on Eq. \ref{eq:aggreg}, taking these two biclusters as the arguments.

Let $B, C, D$, and $E$ be biclusters.
Note that for $D = aggreg(B,C)$, $ov(D,E) \geq ov(B,E)$ and $ov(D,E) \geq ov(C,E)$.
So, for all biclusters $E$ where $ov(B,E) \geq th$ or $ov(C,E) \geq th$, we have $ov(D,E) \geq th$.
For this reason, the order of the aggregation does not interfere on the final result.
It is also important to note that the new bicluster $D$ can have $ov(D,E) \geq th$, for some bicluster $E$ where $ov(B,E) < th$ and $ov(C,E) < th$.
In this aggregation proposal, maximal biclusters will properly merge with non-maximal biclusters.

\subsection{Outlier Removal}
\label{aggregation:outlier}
After aggregating the results, we need to process each final bicluster to look for objects and / or attributes that may be interpreted as outliers.
In this work, this step will always be executed after the aggregation using any of our two proposals.

Let $B = (B^r, B^c)$ be an aggregated bicluster, with $|B^r| = o, |B^c| = p$.
We define a participation matrix $\textbf{P} \in \mathbb{Z}^{o \times p}$, where each element $p_{ij}$ indicates the quantity of biclusters in which this element takes part in $B$.
For example, if an element is part of 15 biclusters that compose $B$, then its value on the $\textbf{P}$ matrix will be 15.

\begin{figure}[]
    \centering
    \includegraphics[trim=2cm 20.5cm 7cm 2.5cm, clip=true, width=200pt]{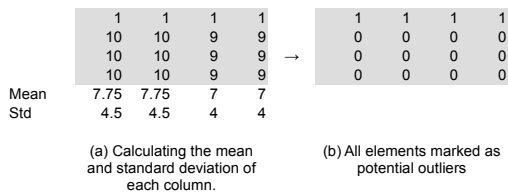}
    \caption{Example of outlier removal.}
    \label{fig:outlier}
\end{figure}

So, we will explain the process of outlier removal with the help of Figure \ref{fig:outlier}.
We have two steps of outlier removal: one for the objects, the other for the attributes.
To remove possibly outlier objects, we take the mean and the standard deviation of all columns on the participation matrix \textbf{P}.
The left side of Figure \ref{fig:outlier} illustrates this step.
After that, we check the values of each element of the columns.
If the value is less than the mean minus one standard deviation, then we check this element as a potential outlier.
In Figure \ref{fig:outlier}, we can see that the entire first row was checked as potential outlier because $1 < 7.75 - 4$.
If we mark the entire row as a potential outlier, it is removed from the bicluster.
In our example, that is the case.

We execute the same process for the columns, calculating the mean, standard deviation and checking for potential outliers on the rows.
We remove the column if it is entirely marked as a potential outlier.

\section{Metrics for Biclustering}
\label{metrics}
In this paper we will use only external metrics, except for the Gene Ontology Enrichment Analysis (GOEA).
External metrics compare a given solution with a reference one.
For an extensive comparison of external metrics for biclustering solutions, the reader may refer to \cite{Horta2014}.

The Gene Ontology Project \footnote{http://geneontology.org} (GO) is an initiative to develop a computational representation of the knowledge of how genes encode biological functions at the molecular, cellular and tissue system levels.
The GOEA compares a set of genes with known information.
For example, given a set of genes that are up-regulated under certain conditions, an enrichment analysis will find which GO terms are over-represented (or under-represented) using annotations for that gene set\footnote{http://geneontology.org/page/about Acessed on 2015, January, 16}.
This method is commonly used to analyze results from biclustering techniques on microarray gene expression datasets.

\textit{Precision}, \textit{Recall} and \textit{F-score} are often used on information retrieval for measuring binary classification \cite{Salton1971}. 
If we take pairs of elements, we can extend these metrics to evaluate clustering / biclustering solutions with overlapping.
The pairwise definition of \textit{Precision} and \textit{Recall} can be found in \cite{Menestrina2009}.
It is important to highlight that these metrics do not consider the quantity of biclusters.
\textit{Pairwise Precision}, or just \textit{Precision} for simplicity, is the fraction of retrieved pairs that are relevant; while \textit{Pairwise Recall}, or just \textit{Recall} for simplicity, is the fraction of relevant pairs that are retrieved.
The \textit{F-score} is the harmonic mean of \textit{Precision} and \textit{Recall}.

Clustering Error (\textit{CE}) is an external metric that considers the quantity of biclusters in its evaluation.
This metric severely penalizes a solution with more biclusters than the reference, thus not being recommended for evaluating enumerative results.
The definition and more details can be found in \cite{Horta2014}.

We propose the difference in coverage, that measures what the reference biclustering solution covers and the found biclustering solution does not cover, and vice versa.
Although very similar, when compared with the pairwise definitions of \textit{Precision} and \textit{Recall}, this metric gives a more intuitive idea of how two solutions cover distinct areas of the dataset.
It also can be computed much faster.
Let $\cup_{\bar{B}} = \bigcup B_i^r \times B_i^c$ be the usual union set of a biclustering solution $\bar{B}$. Let $\bar{B}$ and $\bar{C}$ be the found and the reference biclustering solutions, respectively.
Then the difference in coverage is given by:
\begin{equation}
  dif\_cov(\bar{B}, \bar{C}) = \frac{|\cup_{\bar{B}} - \cup_{\bar{C}}| + |\cup_{\bar{C}} - \cup_{\bar{B}}|}{m \times n}.
  \label{eq:diff_cov}
\end{equation}

We will use this measure to verify how different an aggregated solution is from the enumerative one.

\section{Experiments}
\label{experiments}
In our experiments, we employed three artificial datasets: \textit{art1}, \textit{art2}, and \textit{art3}; and two real datasets: GDS2587 and \textit{FOOD}.
We designed the artificial datasets to present different scenarios with increasing difficulty.
They have 1000 objects and 15 attributes. Each entry is a random integer, drawn from a discrete uniform distribution on the set \{1, 2, ..., 100\}.
Then we inserted: 5 bicluster arbitrarily positioned and without overlapping on \textit{art1}; 5 bicluster arbitrarily positioned and with a similar degree of overlapping on \textit{art2}; and 15 bicluster arbitrarily positioned and with different degrees of overlapping on \textit{art3}.

For each bicluster, the quantity of objects was randomly drawn from the set $\{50, \dots, 60\}$, and the quantity of attributes was randomly drawn from the set $\{4, 5, 6, 7\}$.
To insert a bicluster, we fixed the value of the first attribute and obtained the values of the other attributes by adding a constant value to the first column.
This characterizes biclusters of coherent values.
This constant value was randomly drawn from the set $\{-10, -9, \dots, -1, 1, \dots, 9, 10\}$.

\textit{GDS2587}\footnote{http://www.ncbi.nlm.nih.gov/sites/GDSbrowser?acc=GDS2587} is a microarray gene expression dataset, with 2792 genes and 7 samples, collected from the organism \textit{E. coli}.
We removed every gene with missing data in any sample, and the data was normalized by mean centralization, as usual in gene expression data analysis \cite{Prelic2006}.
In this dataset we aim to validate our contribution when devoted to microarray gene expression data analysis, as it is considered a relevant application of biclustering methods.

\textit{FOOD}\footnote{http://www.ntwrks.com/chart1a.htm} is a dataset with 961 objects, which represent different foods, and 7 attributes, which represent nutritional information.
As the values of each attribute are in different ranges, we used the same pre-processing as Veroneze \textit{et al}. \cite{Veroneze2014}.
In this dataset our goal is to illustrate the usefulness of bicluster aggregation in a different scenario and to verify if the aggregation leaves uncovered areas that the enumeration has covered at first.

\subsection{Experiments on Artificial Datasets}
Our goal is to verify the impact of noise in the enumeration of biclusters, and how the aggregation can improve the quality of the final results.
To this end, we will add a Gaussian noise with $\mu = 0$ and $\sigma \in \{0, 0.01, \dots, 1\}$, to each dataset, and then run the RIn-Close algorithm.
This procedure will be repeated for 30 times and all reported values will be the average of this 30 executions.
We will set RIn-Close to mine coherent values biclusters, with at least 50 rows and 4 columns.
Also, we will use crescent values for $\epsilon$ due to the importance of the parameter.
If $\epsilon$ is too small, we may miss important biclusters expressing more internal variance.
If $\epsilon$ is too high, the biclusters may include unexpected objects or attributes.

As we know the biclusters, we will use \textit{Precision}, \textit{Recall} and \textit{F-score} to assess the quality of the results after the enumeration.
After that, we will perform the aggregation on the results with the value of $\epsilon$ that led to an initial \textit{Precision} closest to 0.85.
This value was chosen because if the \textit{Precision} is too low, it means that the $\epsilon$ value is allowing too many undesired objects or attributes in the enumerated biclusters.
In this case, the aggregation may not improve the quality of the final results because their input is not of good quality.
If the \textit{Precision} is too high, we will only be able to see improvements in the reduced quantity of biclusters, but the aggregation may increase the \textit{Precision} too.

We will consider the following algorithms as contenders:
\begin{itemize}
\item[Triclustering \cite{Hanczar2012}.]
We set $k$ to the true number of biclusters. The authors supplied the code for this algorithm.

\item[Merging and Deleting steps of MicroCluster \cite{Zhao2005}.]
To parameterize this algorithm, we ran a grid search with the values in the set ${0.15, 0.1, 0.05}$, getting 9 results for each run.
Also, as the aggregation step of the algorithm is composed of two steps, merging and deleting, we ran each experiment twice: with the merging step first (MD) and with the deleting step first (DM).
Unless we want to draw attention to some particular fact, we will report only the best result.
The authors supplied the code for this algorithm \footnote{http://www.cs.rpi.edu/$\sim$zaki/www-new/pmwiki.php/Software/Software}.

\item[Single Linkage (see Section \ref{aggregation:sl}).]
We cut the dendrogram with the proper quantity of biclusters: for \textit{art1} and \textit{art2}, 5 biclusters; for \textit{art3}, 15 biclusters.

\item[Aggregation by Overlapping (see Section \ref{aggregation:by_ov}).]
We tested several values for the rate of overlapping.

\end{itemize}
After getting the results for all executions of the listed algorithms, we will choose the best result from each one and compare them using the \textit{CE} metric.

\begin{figure}
    \centering
    \begin{subfloat}[\textit{art1}]{
        \includegraphics[trim=2.4cm 7.5cm 3.5cm 8.3cm, clip=true, width=100pt]{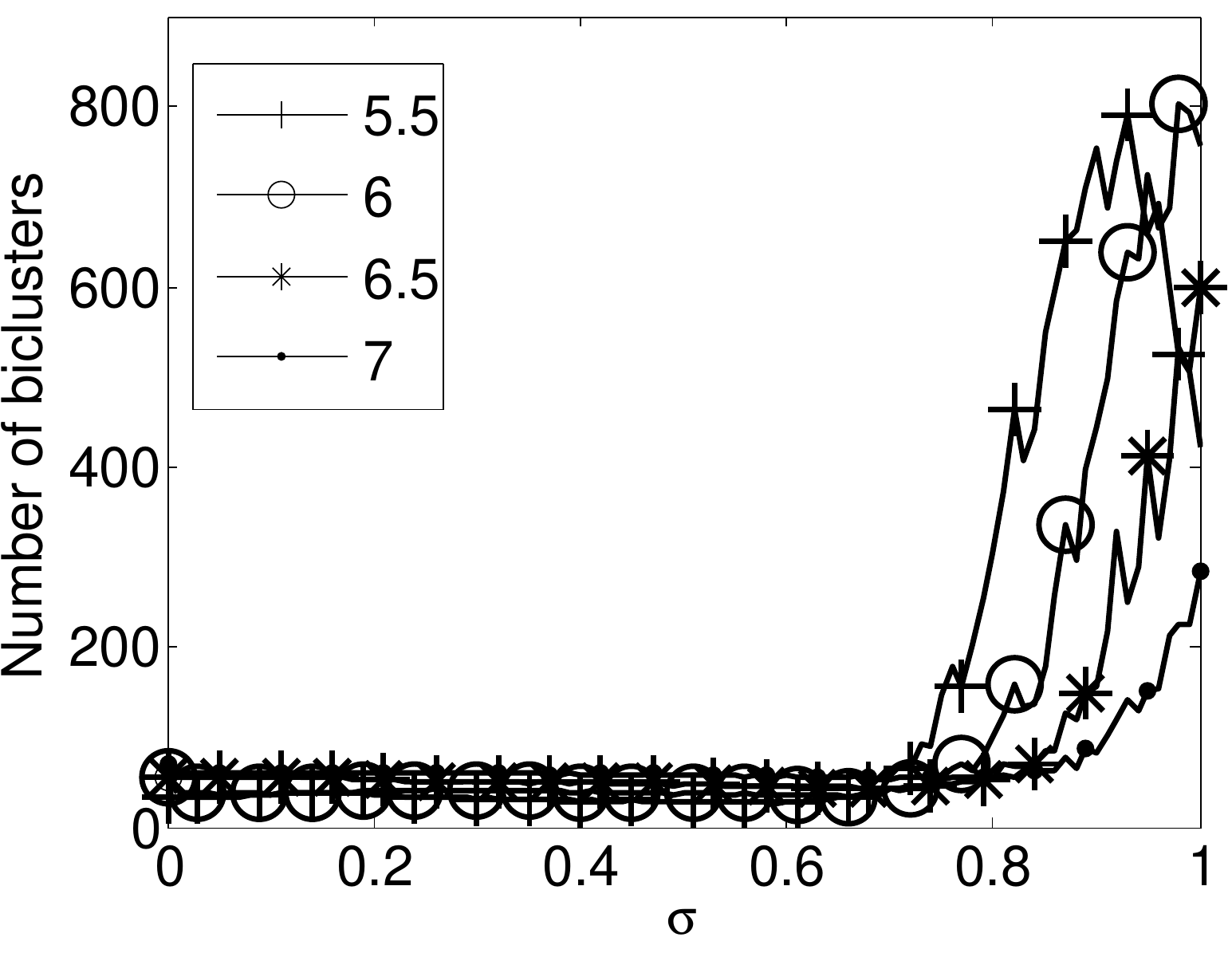} 
        \label{fig:qtds:art1}
    }
    \end{subfloat}
    \begin{subfloat}[\textit{art2}]{
        \includegraphics[trim=2.4cm 7.5cm 3.5cm 8.3cm, clip=true, width=100pt]{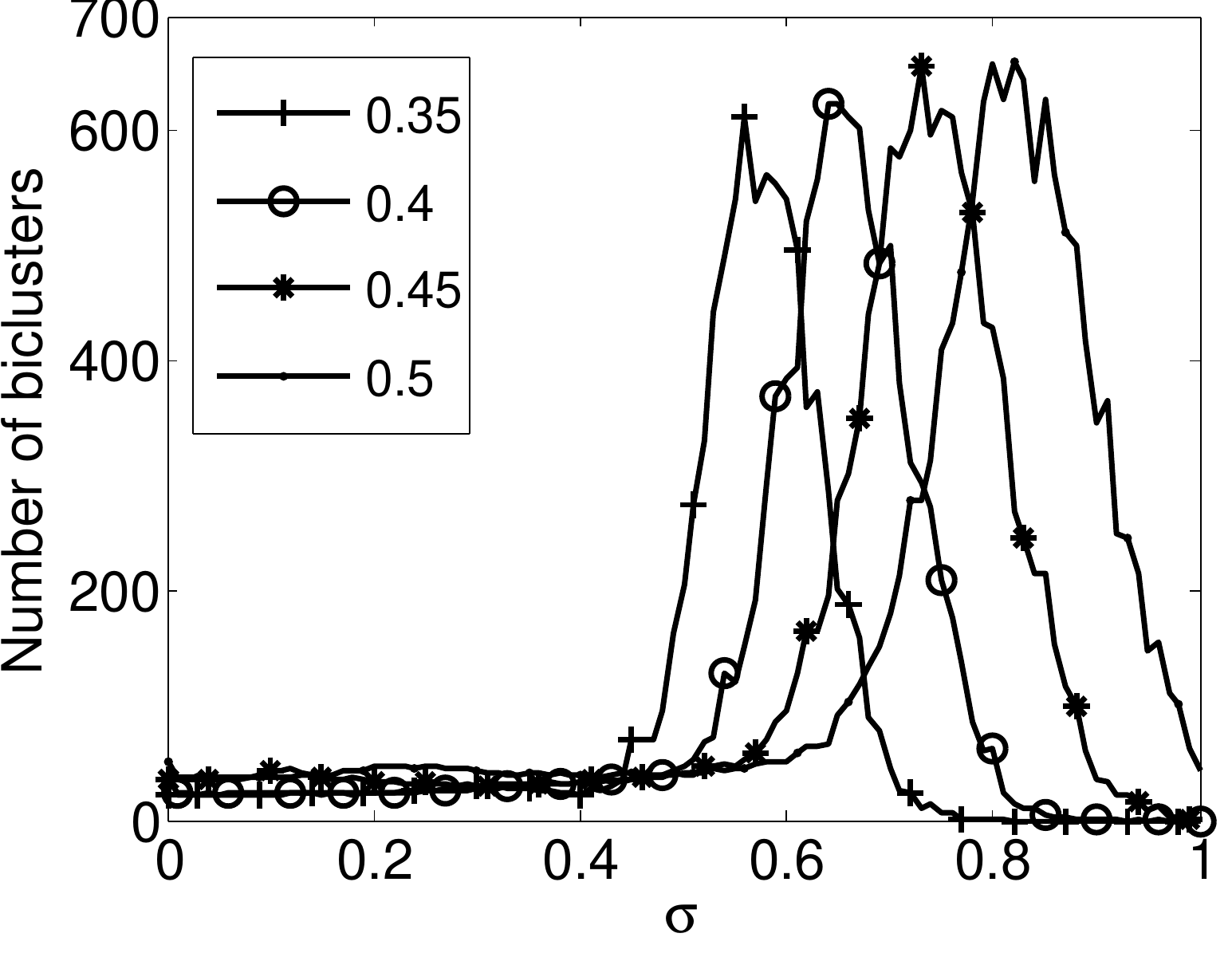}
        \label{fig:qtds:art2}
    }
    \end{subfloat}
    \begin{subfloat}[\textit{art3}]{
        \includegraphics[trim=2.4cm 7.5cm 3.5cm 8.3cm, clip=true, width=100pt]{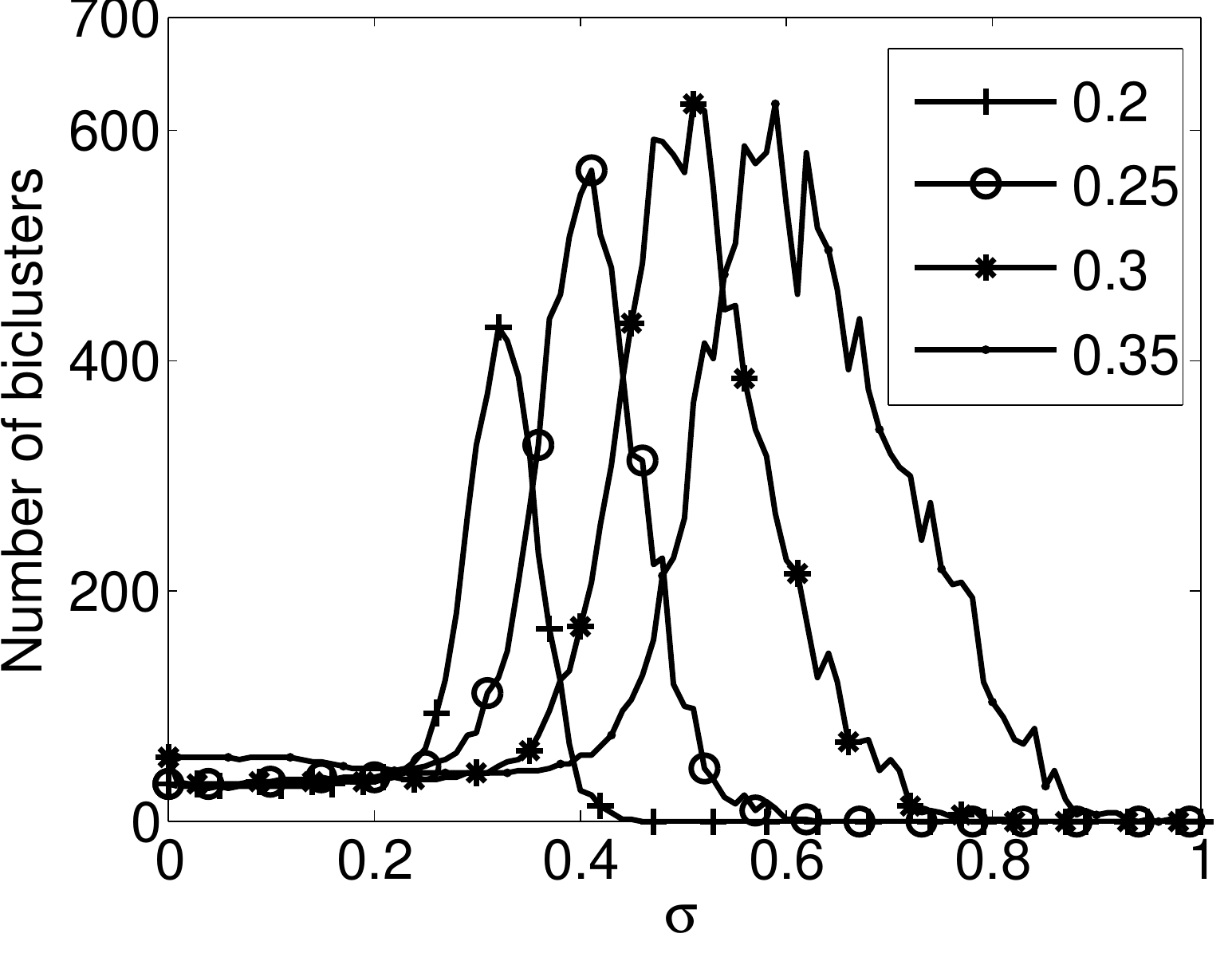}
        \label{fig:qtds:art3}
    }
    \end{subfloat}
    \caption{Quantity of enumerated biclusters by the variance of the Gaussian noise in the artificial datasets.
    Each curve is parameterized by $\epsilon$.}
    \label{fig:qtds}
\end{figure}

Figure \ref{fig:qtds} shows the quantity of enumerated biclusters on the artificial datasets, for several values of $\epsilon$.
In all datasets, for every value of $\epsilon$, the behavior is the same: as the noise increases the quantity of enumerated biclusters starts to increase.
In Figures \ref{fig:qtds:art1} and \ref{fig:qtds:art2}, we know that the real quantity of biclusters is 5, but when the noise increases, the enumerated quantity reaches approximately 800 biclusters, depending on the value of $\epsilon$.
In Figure \ref{fig:qtds:art3}, we can see that the quantity of biclusters reaches high values too.
At some level of noise, the number of biclusters starts to decrease to a point that the algorithm is not able to find any bicluster.

\begin{figure}[h]
    \centering
    \begin{subfloat}[\textit{art1} \textit{Precision}]{
        \includegraphics[trim=3.1cm 7.5cm 3.5cm 8.5cm, clip=true, width=100pt]{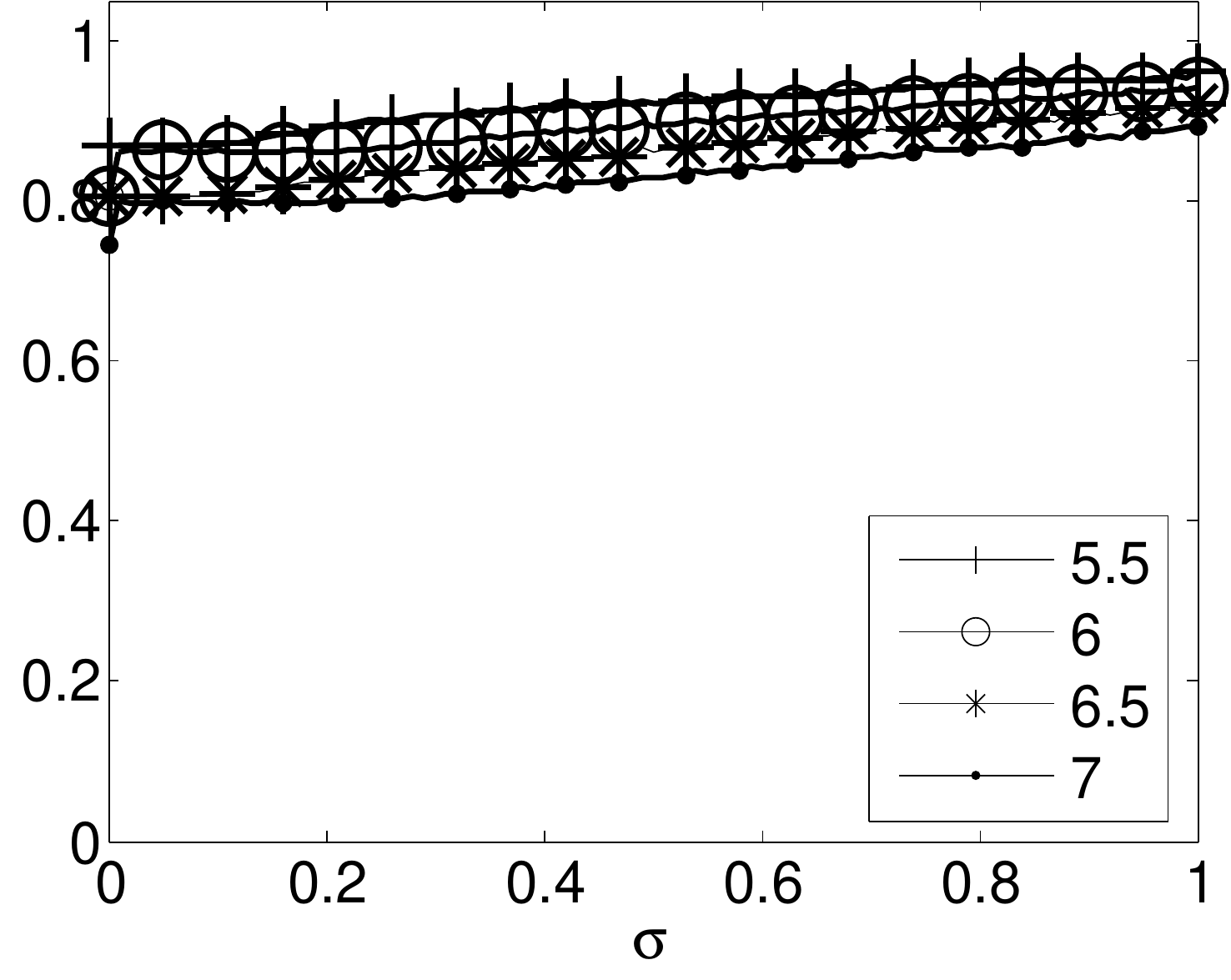} 
        \label{fig:art1_precision}
    }
    \end{subfloat}
    \begin{subfloat}[\textit{art2} \textit{Precision}]{
        \includegraphics[trim=3.1cm 7.5cm 3.5cm 8.5cm, clip=true, width=100pt]{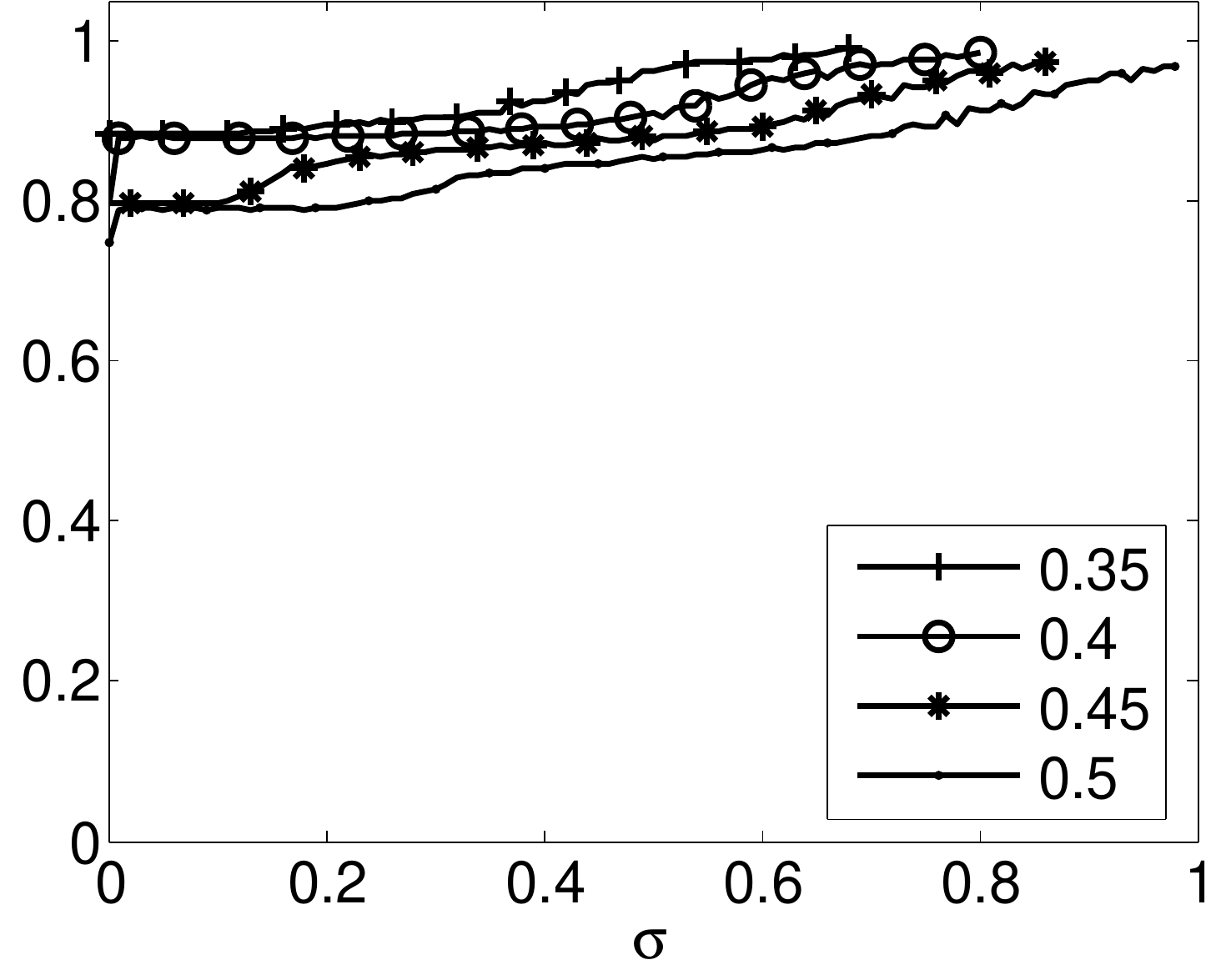} 
        \label{fig:art2_precision}
    }
    \end{subfloat}
    \begin{subfloat}[\textit{art3} \textit{Precision}]{
        \includegraphics[trim=3.1cm 7.5cm 3.5cm 8.5cm, clip=true, width=100pt]{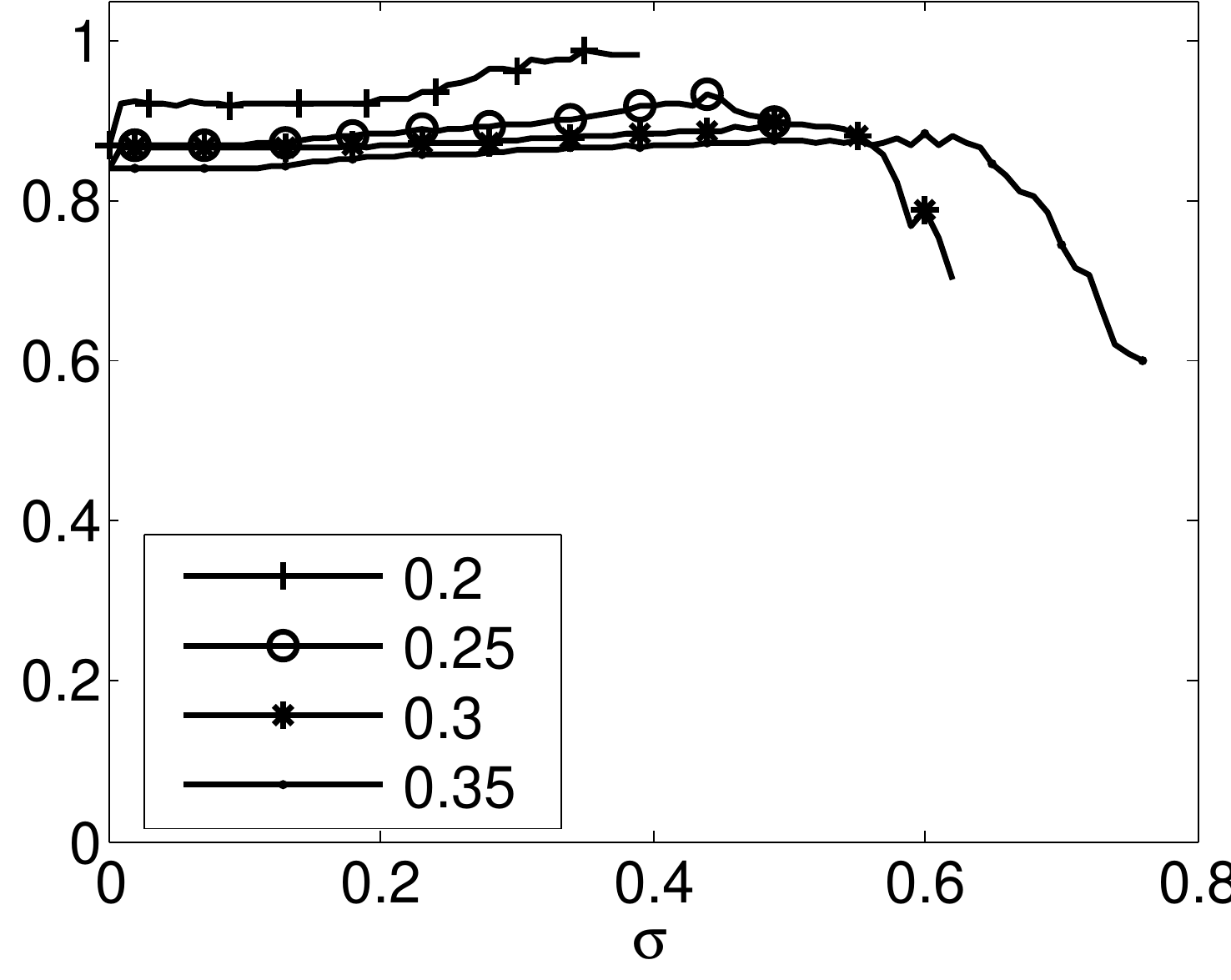} 
        \label{fig:art3_precision}
    }
    \end{subfloat}
    \begin{subfloat}[\textit{art1} \textit{Recall}]{
        \includegraphics[trim=3.1cm 7.5cm 3.5cm 8.5cm, clip=true, width=100pt]{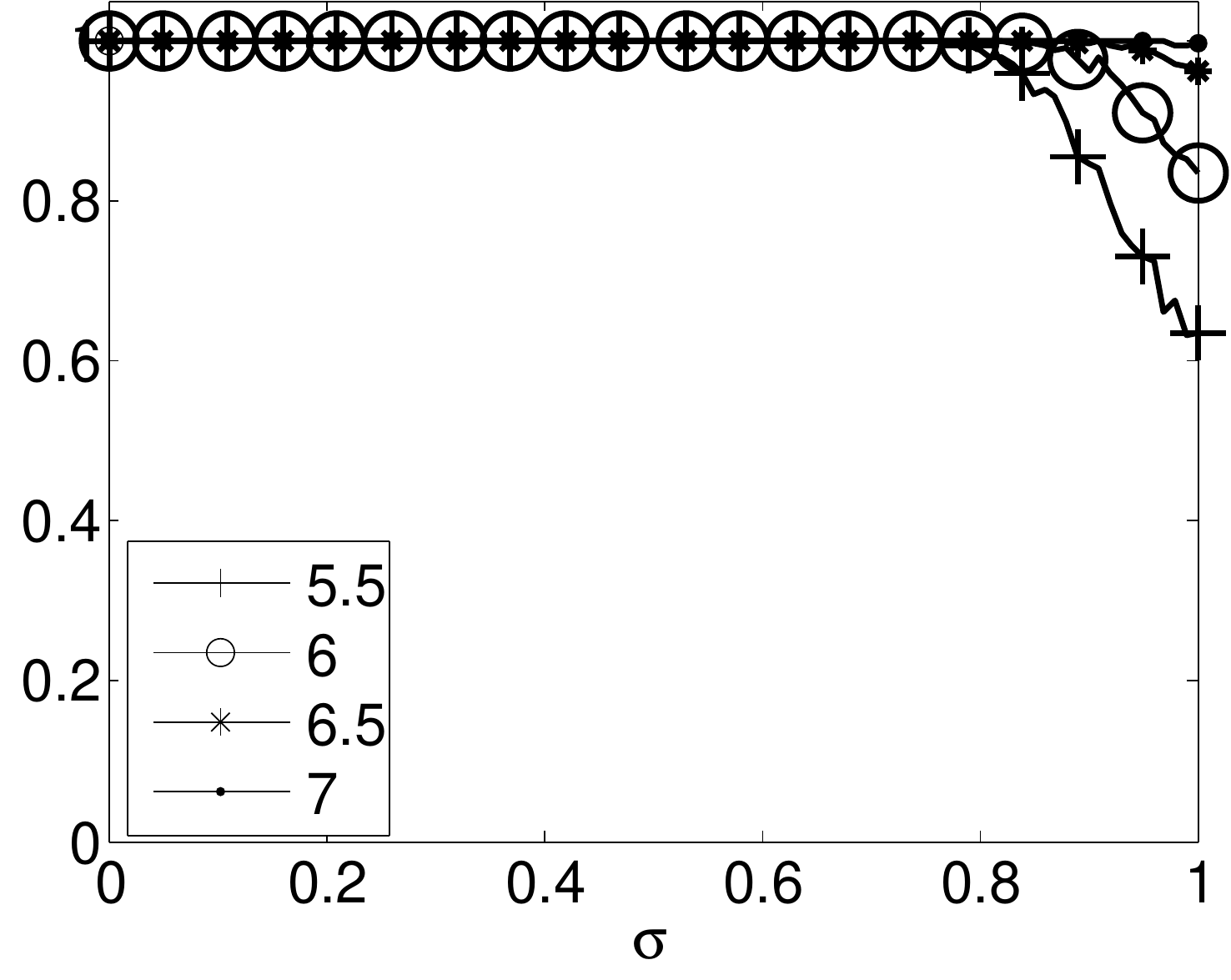} 
        \label{fig:art1_recall}
    }
    \end{subfloat}
    \begin{subfloat}[\textit{art2} \textit{Recall}]{
        \includegraphics[trim=3.1cm 7.5cm 3.5cm 8.5cm, clip=true, width=100pt]{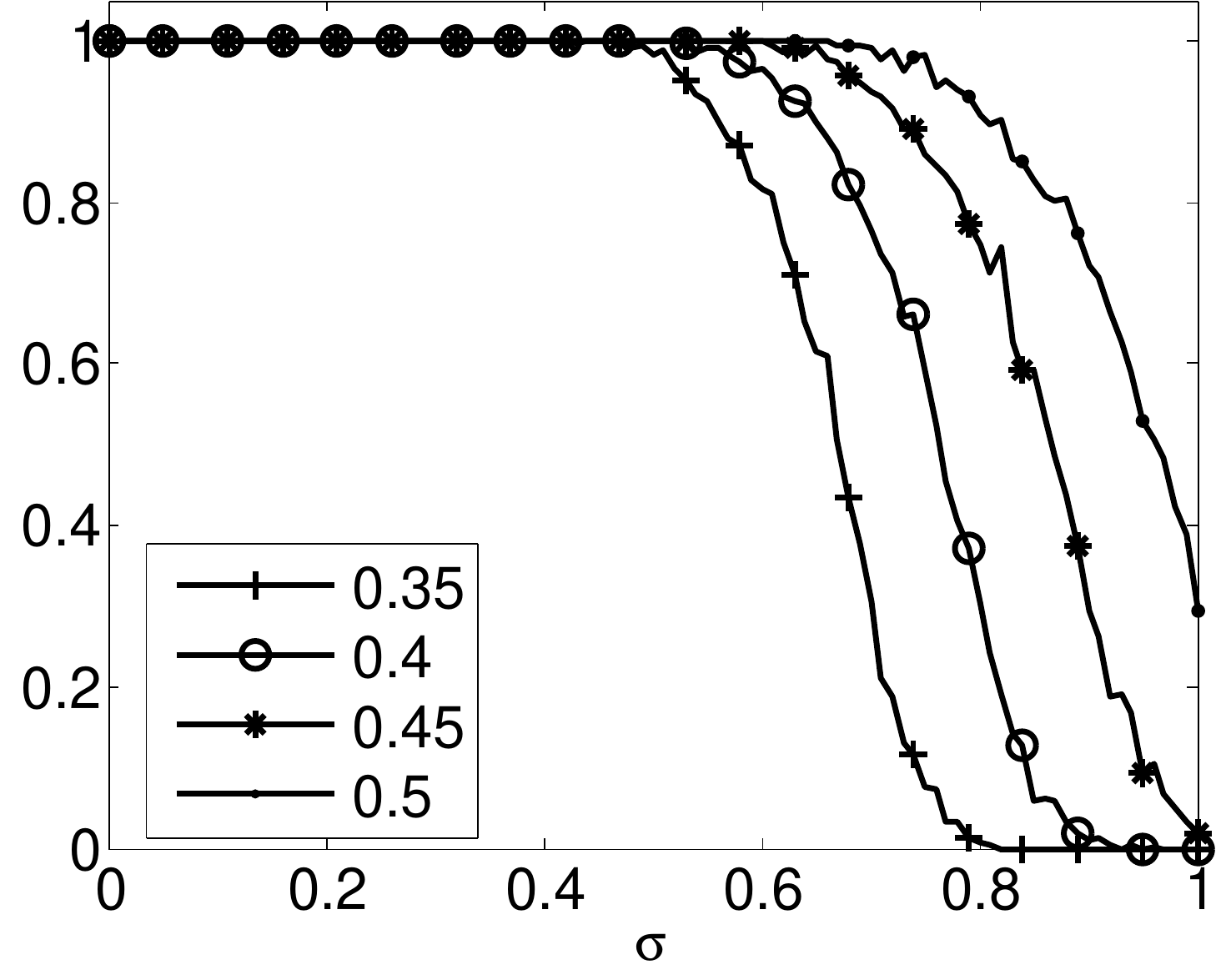} 
        \label{fig:art2_recall}
    }
    \end{subfloat}
    \begin{subfloat}[\textit{art3} \textit{Recall}]{
        \includegraphics[trim=3.1cm 7.5cm 3.5cm 8.5cm, clip=true, width=100pt]{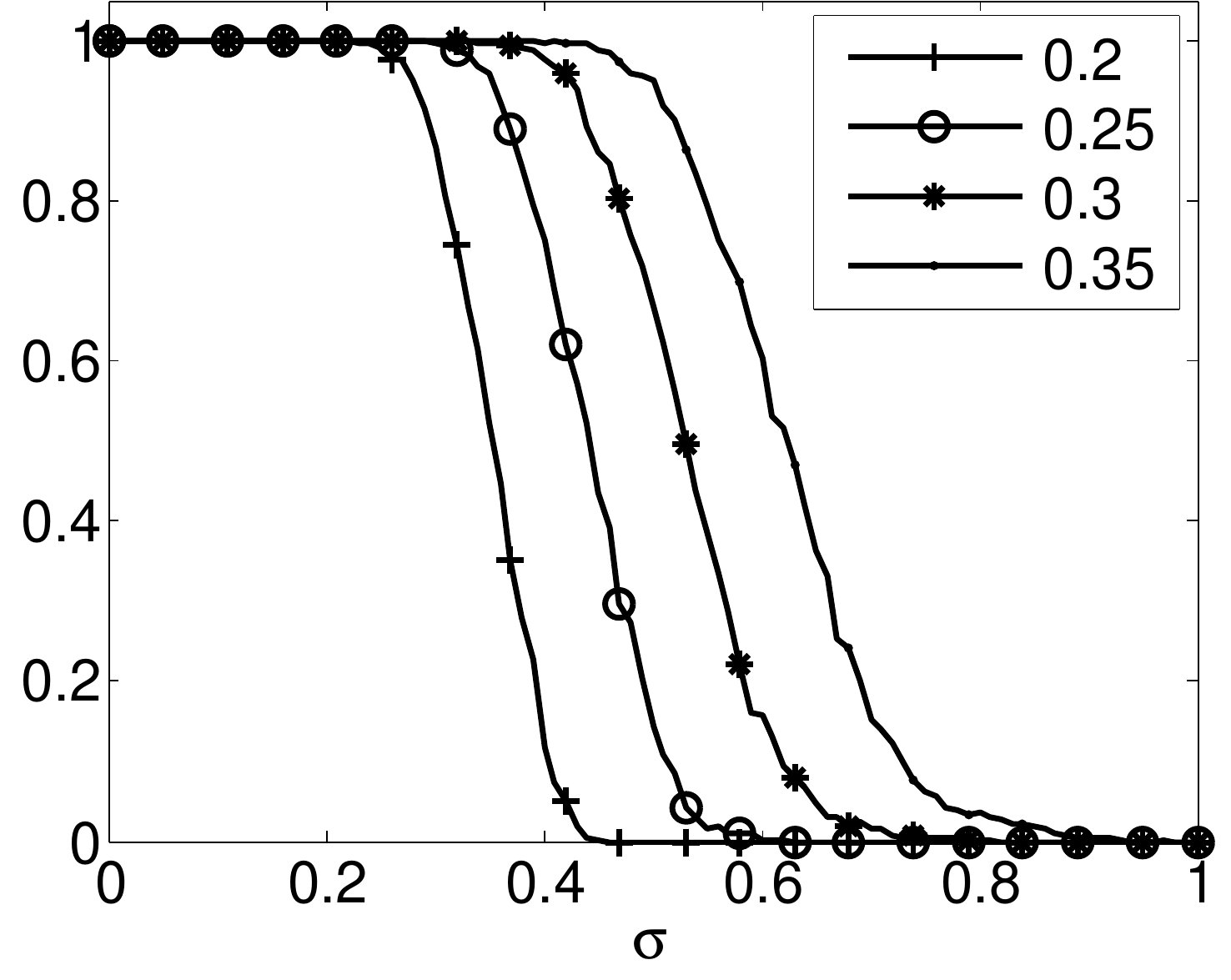} 
        \label{fig:art3_recall}
    }
    \end{subfloat}
    \caption{\textit{Precision} and \textit{Recall} for the solutions of RIn-Close, with several values of $\epsilon$, by the variance of the Gaussian noise in the artificial datasets.}
    \label{fig:prec_rec}
\end{figure}
In Figure \ref{fig:prec_rec}, we can see the quality of the enumeration without considering the quantity of biclusters.

As we can see in Figure \ref{fig:art1_recall}, the noise has almost no interference in the recall for \textit{art1}.
It means that this dataset has biclusters very well defined, that even with some noise they are not missed.
On the other hand, when the variance of the noise is too low, Figure \ref{fig:art1_precision} shows that the found biclusters contains more elements than expected.
It is happening because the parameter $\epsilon$ is high, allowing some elements to be part of the biclusters even without being part of the original solution.
As the noise increases, less of these intruder elements are going to satisfy the $\epsilon$ restriction to be thus included in some bicluster.
In this dataset, the effect of the noise were not so severe on the quality, given that the recall started to decrease only when the variance of the noise was close to 1.

In dataset \textit{art2} the effect of noise can be better observed.
Figure \ref{fig:art2_recall} shows that the noise starts to affect the solutions very early.
When $\epsilon = 3$, the recall starts to decrease very soon, with $\sigma \approx 0.5$.
However, for more relaxed values of $\epsilon$ we can still see the decrease on the recall.
Being the most difficult, dataset \textit{art3} is the most affected by noise.
Independently of the value of $\epsilon$, the RIn-Close was not able to find any biclusters after some levels of variance in the noise.
For example, when $\epsilon = 2$, after $\sigma \approx 0.4$ the \textit{Precision} gets undefined.
This happens because the metric is not defined when the quantity of biclusters is zero.
In Figure \ref{fig:art3_recall}, we can see that the decline of the recall starts when $\sigma \approx 0.3$ for $\epsilon = 2$.

Now we will discuss the results of the aggregation with the previously listed algorithms.
As stated earlier, we will use the results from a value of $\epsilon$ that led to an initial \textit{Precision} close to 0.85.
In this case, we have $\epsilon = 6, 4, 3$ for \textit{art1}, \textit{art2} and \textit{art3}, respectively.
\begin{figure}[]
    \centering
    \begin{subfloat}[\textit{Single Linkage}]{
        \includegraphics[trim=1.6cm 7cm 2.7cm 6.9cm, clip=true, width=100pt]{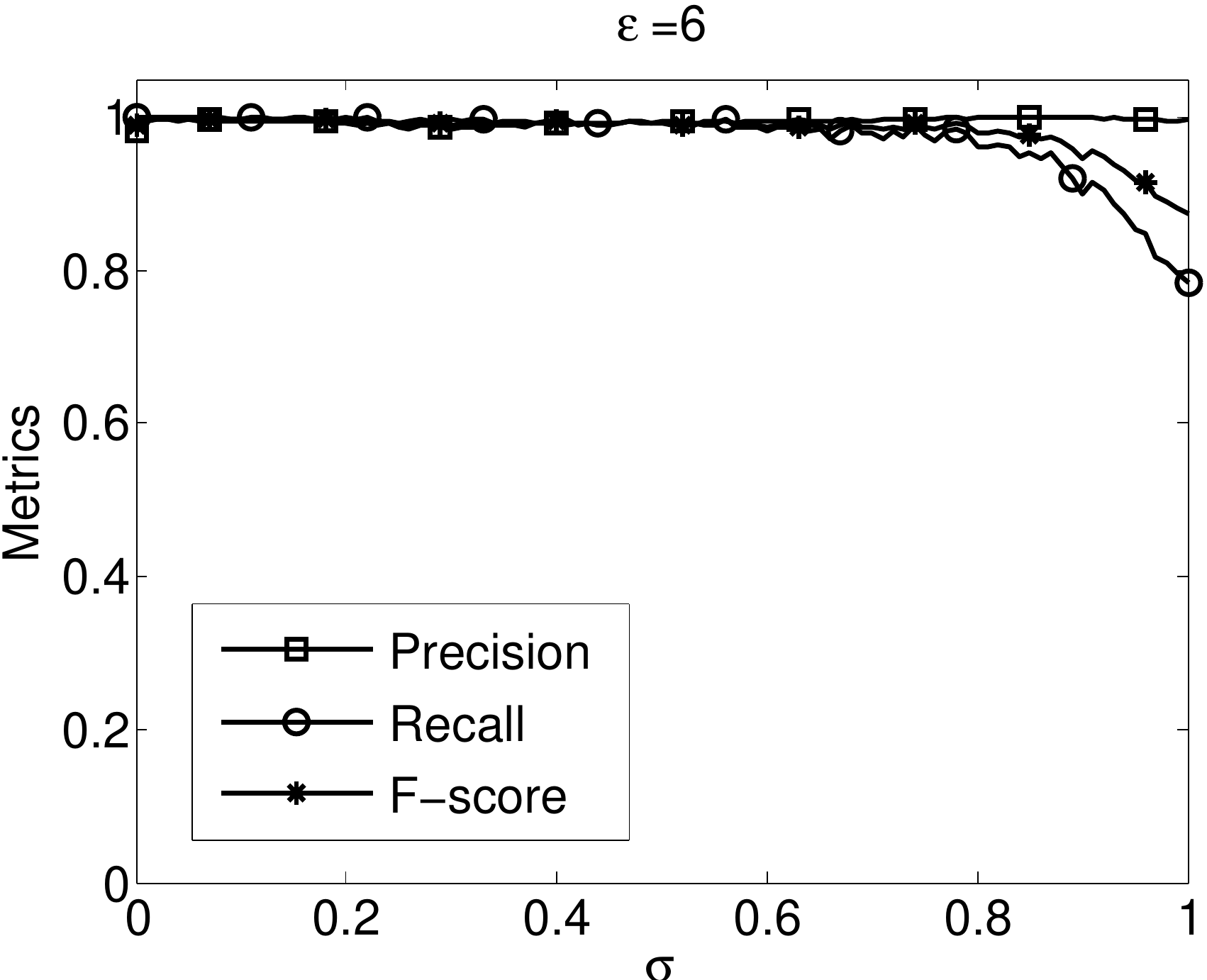}
        \label{fig:aggreg_art1:sl}
    }
    \end{subfloat}
    \begin{subfloat}[\textit{By Overlapping}]{
        \includegraphics[trim=1.4cm 6.5cm 0.7cm 7.1cm, clip=true, width=100pt]{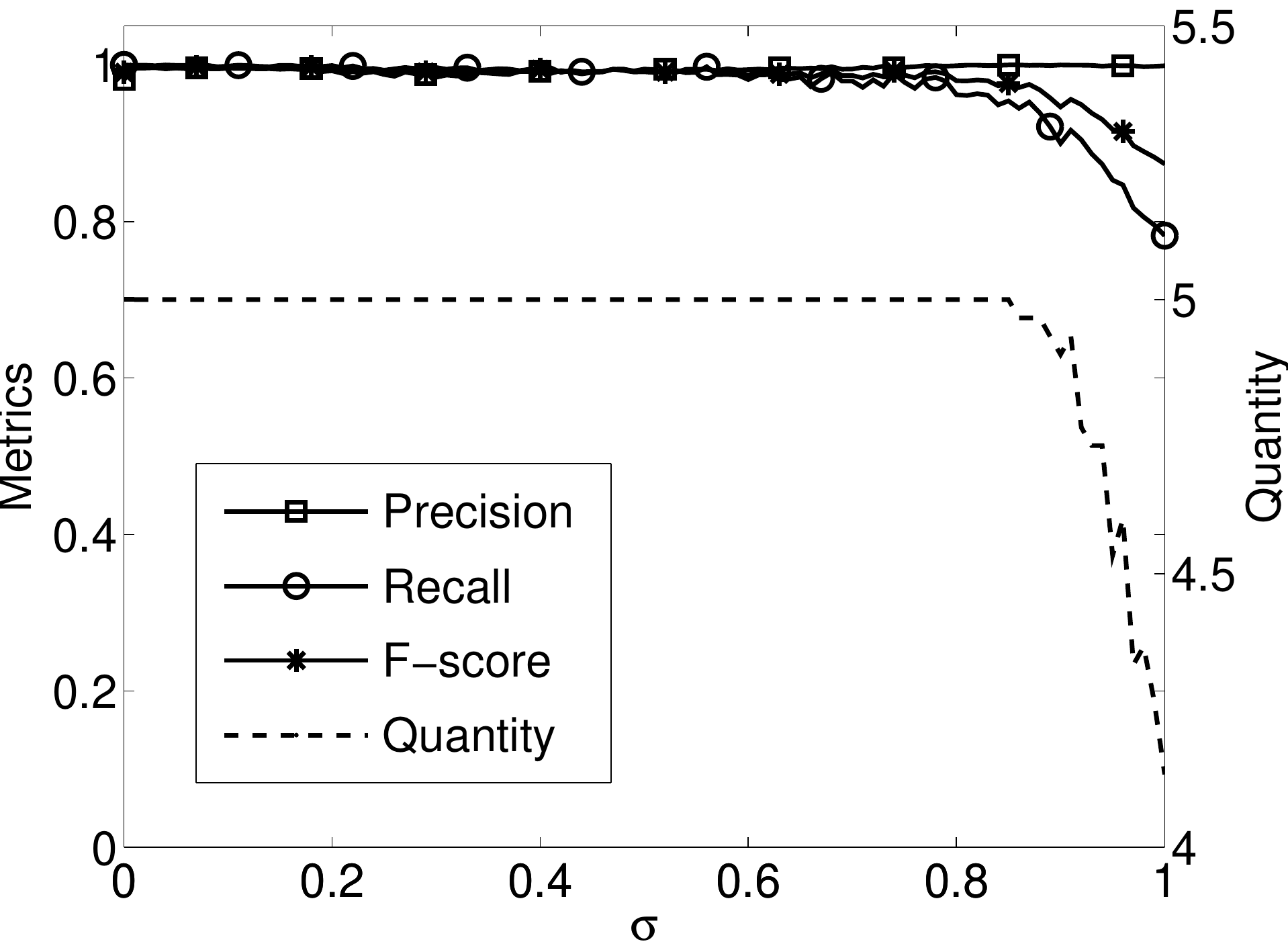}
        \label{fig:aggreg_art1:ov}
    }
    \end{subfloat}
    \begin{subfloat}[\textit{CE}]{
        \includegraphics[trim=2.4cm 7.7cm 2.7cm 7.5cm, clip=true, width=100pt]{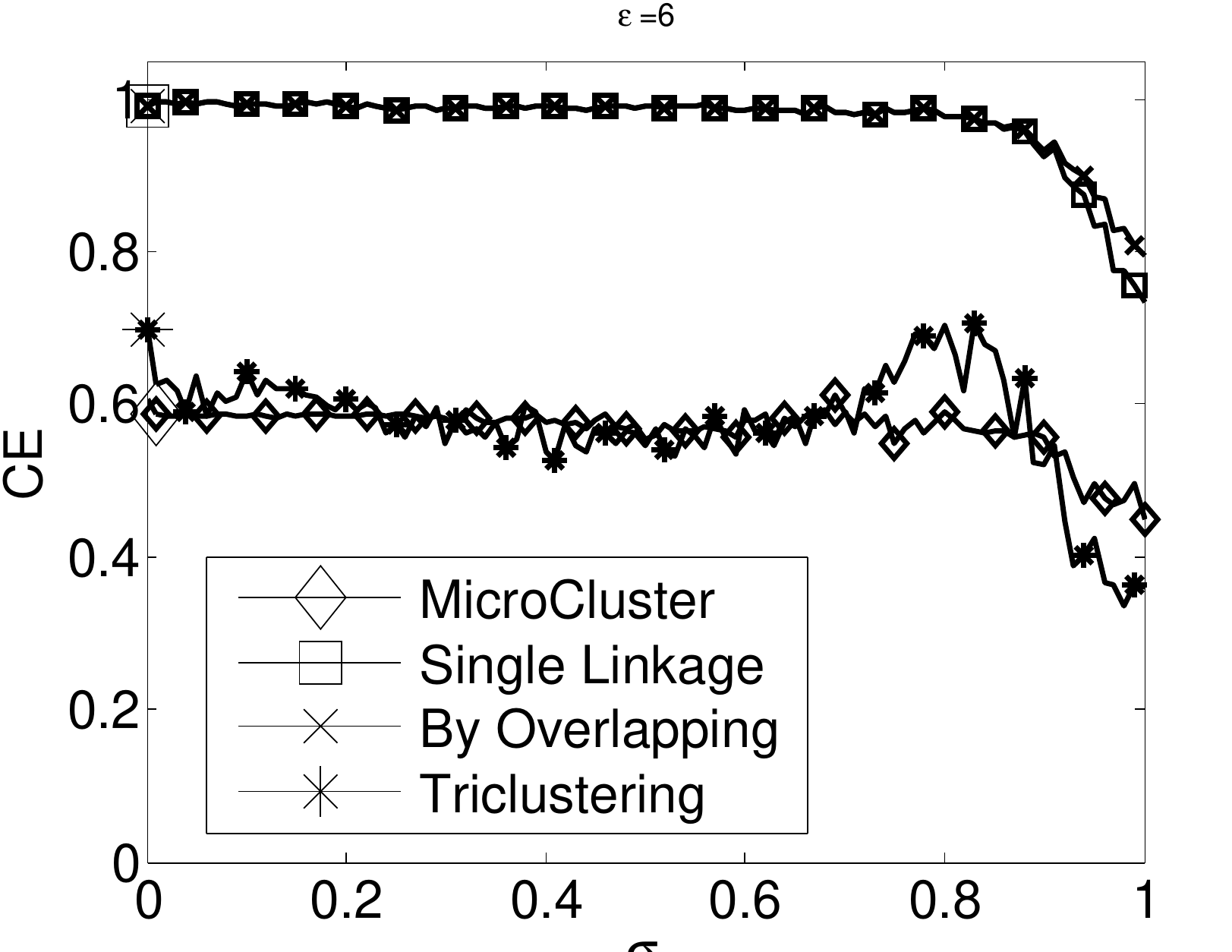}
        \label{fig:aggreg_art1:ces}
    }
    \end{subfloat}
    \caption{Solutions of aggregation as a function of the variance of the noise in dataset \textit{art1}.
    The scale on the right refers to quantity.}
    \label{fig:aggreg_art1}
\end{figure}

Figure \ref{fig:aggreg_art1:sl} shows the quality of the aggregation with single linkage for dataset \textit{art1}.
We can see that, with the proper number of biclusters, the aggregation was able to get an almost perfect result.
The same thing happened with the aggregation by overlapping, reported in Figure \ref{fig:aggreg_art1:ov}.
Figure \ref{fig:aggreg_art1:ces} shows the CE metric for all solutions of aggregation.
We can see that our proposals were capable of producing the best performance on this dataset.

\begin{figure}[]
    \centering
    \begin{subfloat}[\textit{Single Linkage}]{
        \includegraphics[trim=2.9cm 7.7cm 2.7cm 7.5cm, clip=true, width=100pt]{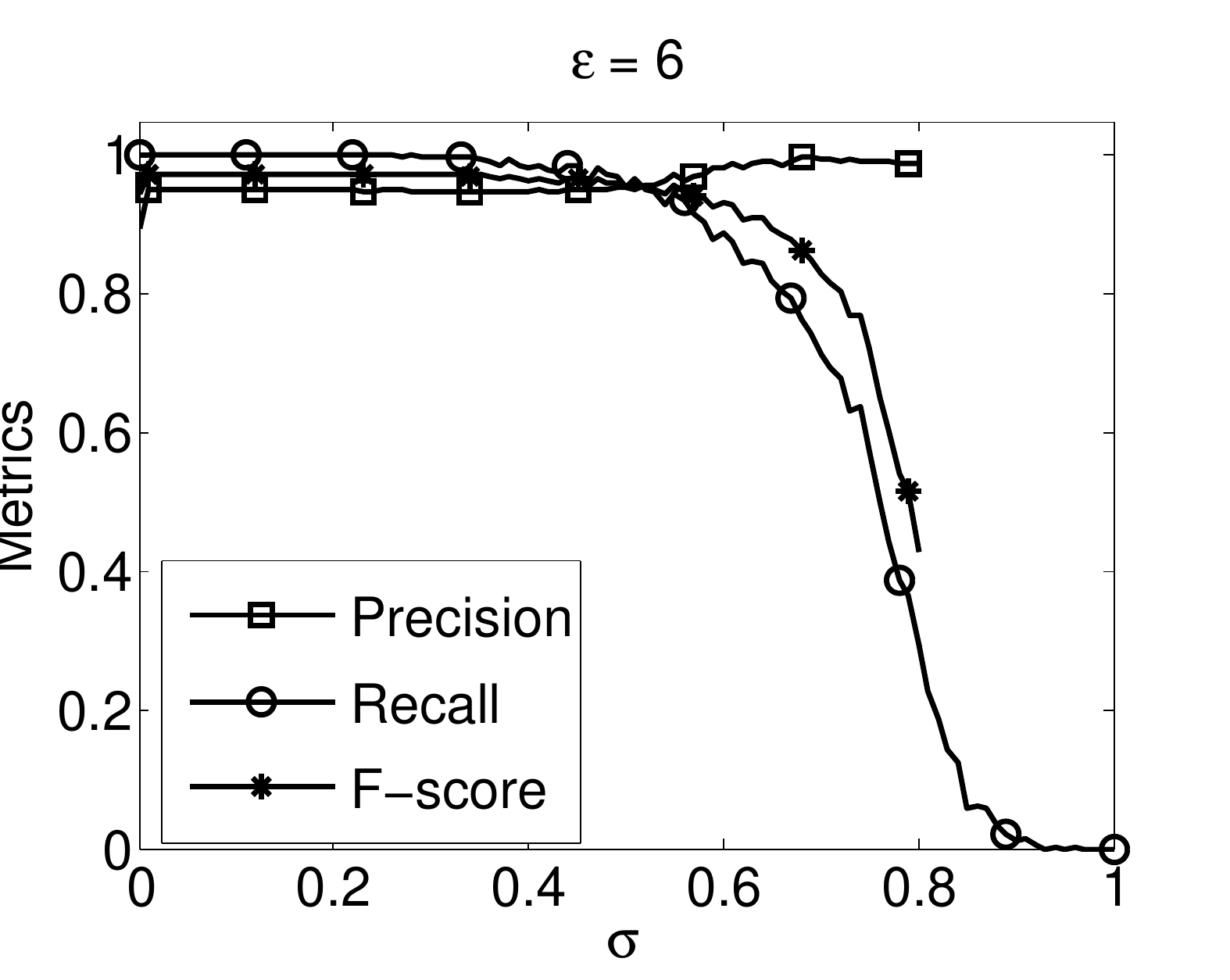}
        \label{fig:aggreg_art2:sl}
    }
    \end{subfloat}
    \begin{subfloat}[\textit{By Overlapping}]{
        \includegraphics[trim=1.7cm 7cm 1.5cm 7.5cm, clip=true, width=100pt]{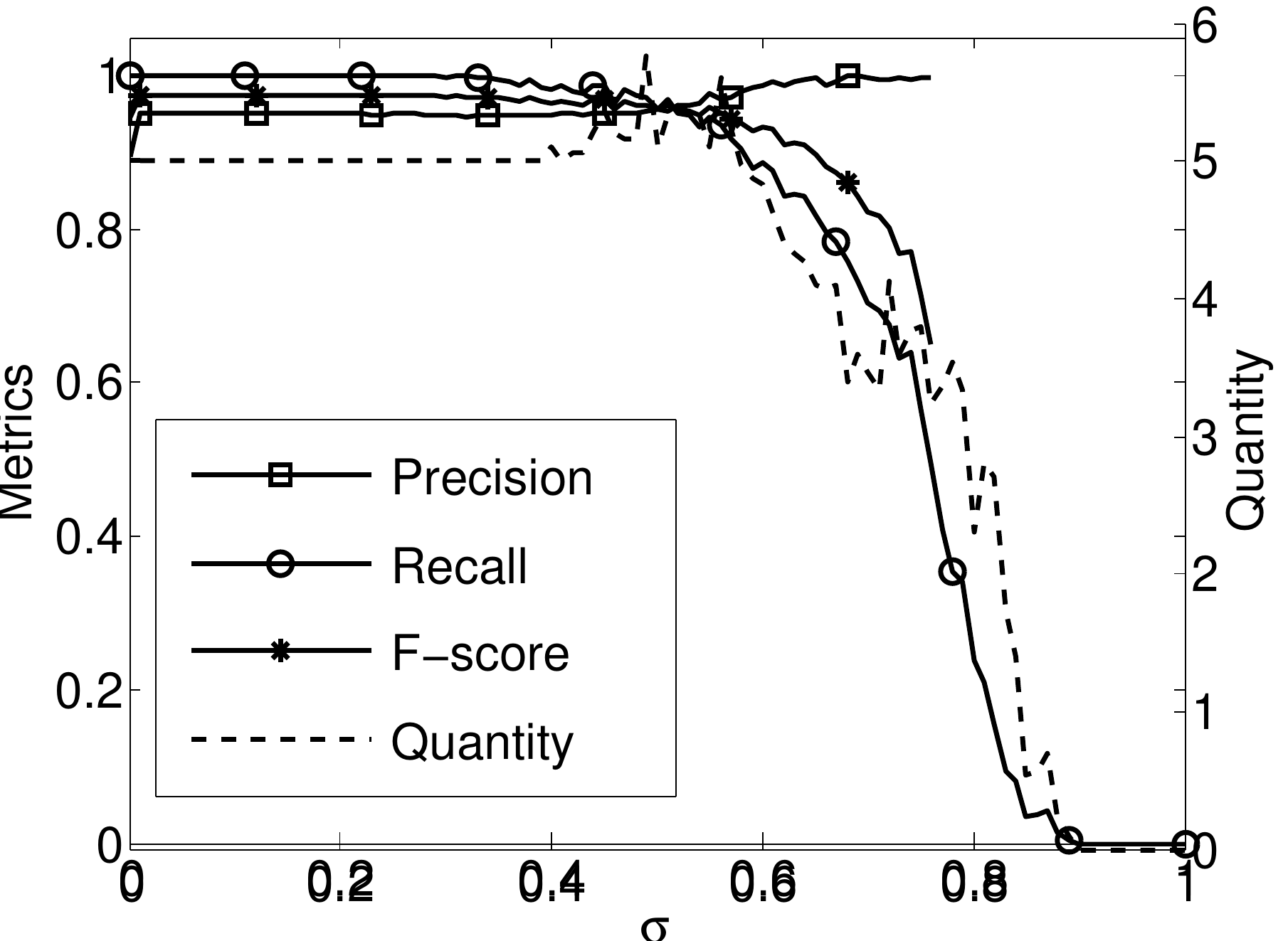}
        \label{fig:aggreg_art2:ov}
    }
    \end{subfloat}
    \begin{subfloat}[\textit{CE}]{
        \includegraphics[trim=2.8cm 7.7cm 2.7cm 7.5cm, clip=true, width=100pt]{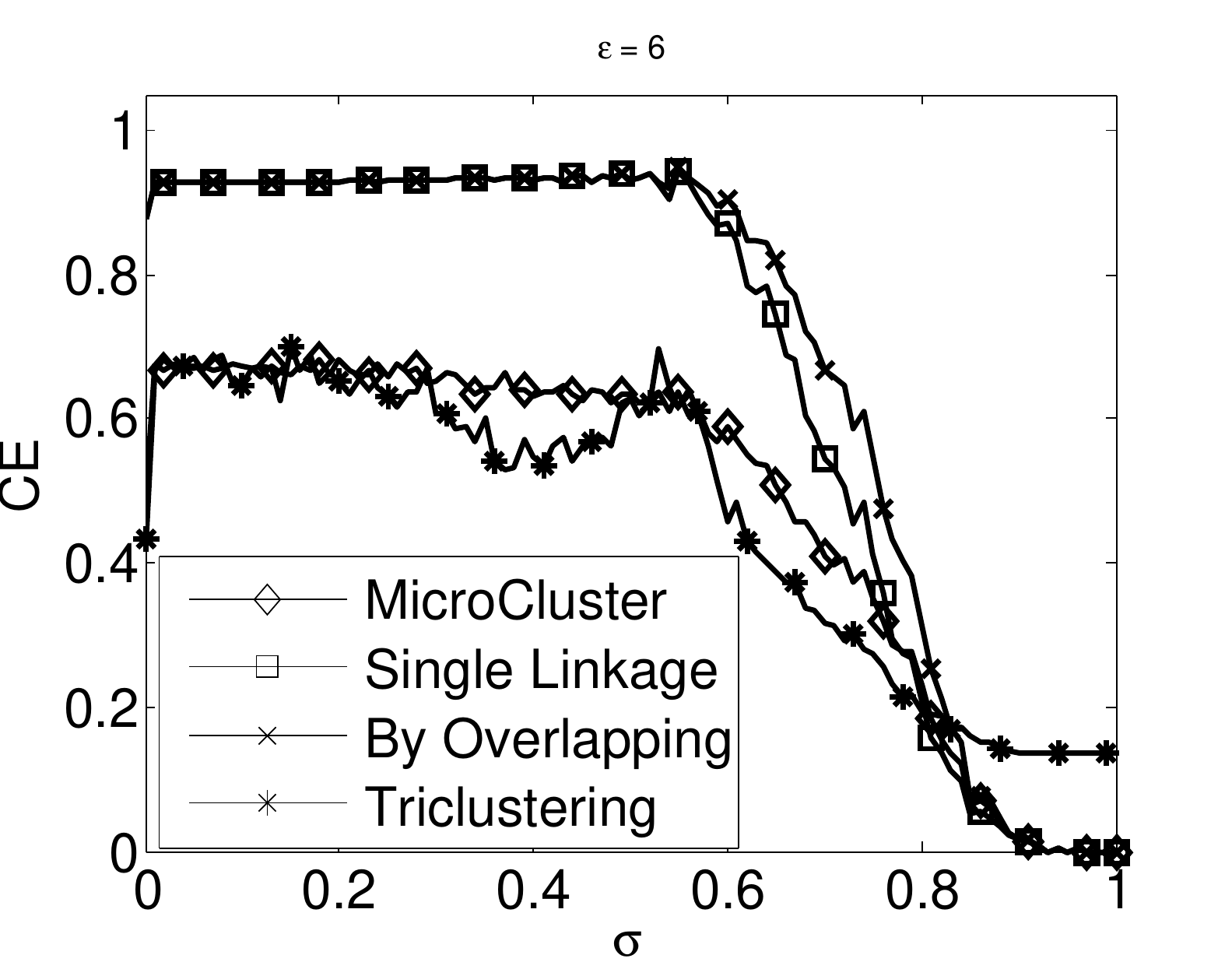}
        \label{fig:aggreg_art2:ces}
    }
    \end{subfloat}
    \caption{Solutions of aggregation as a function of the variance of the noise in dataset \textit{art2}.
    The scale on the right refers to quantity.}
    \label{fig:aggreg_art2}
\end{figure}
    
Figure \ref{fig:aggreg_art2:sl} shows the quality of the aggregation with single linkage for the dataset \textit{art2}.
This time, the solution was close to the maximum achievable performance, but not so close as it was in \textit{art1}.
Figure \ref{fig:aggreg_art2:ov} shows the quality of the aggregation by overlapping for the same dataset.
The quality of this solution is very similar to the one obtained with single linkage.
Figure \ref{fig:aggreg_art2:ces} shows the CE metrics obtained by all the methods of aggregation.
Again, our proposals outperformed the other two algorithms.

\begin{figure}[]
    \centering
    \begin{subfloat}[\textit{Single Linkage}]{
        \includegraphics[trim=2.8cm 8cm 4.1cm 7.9cm, clip=true, width=100pt]{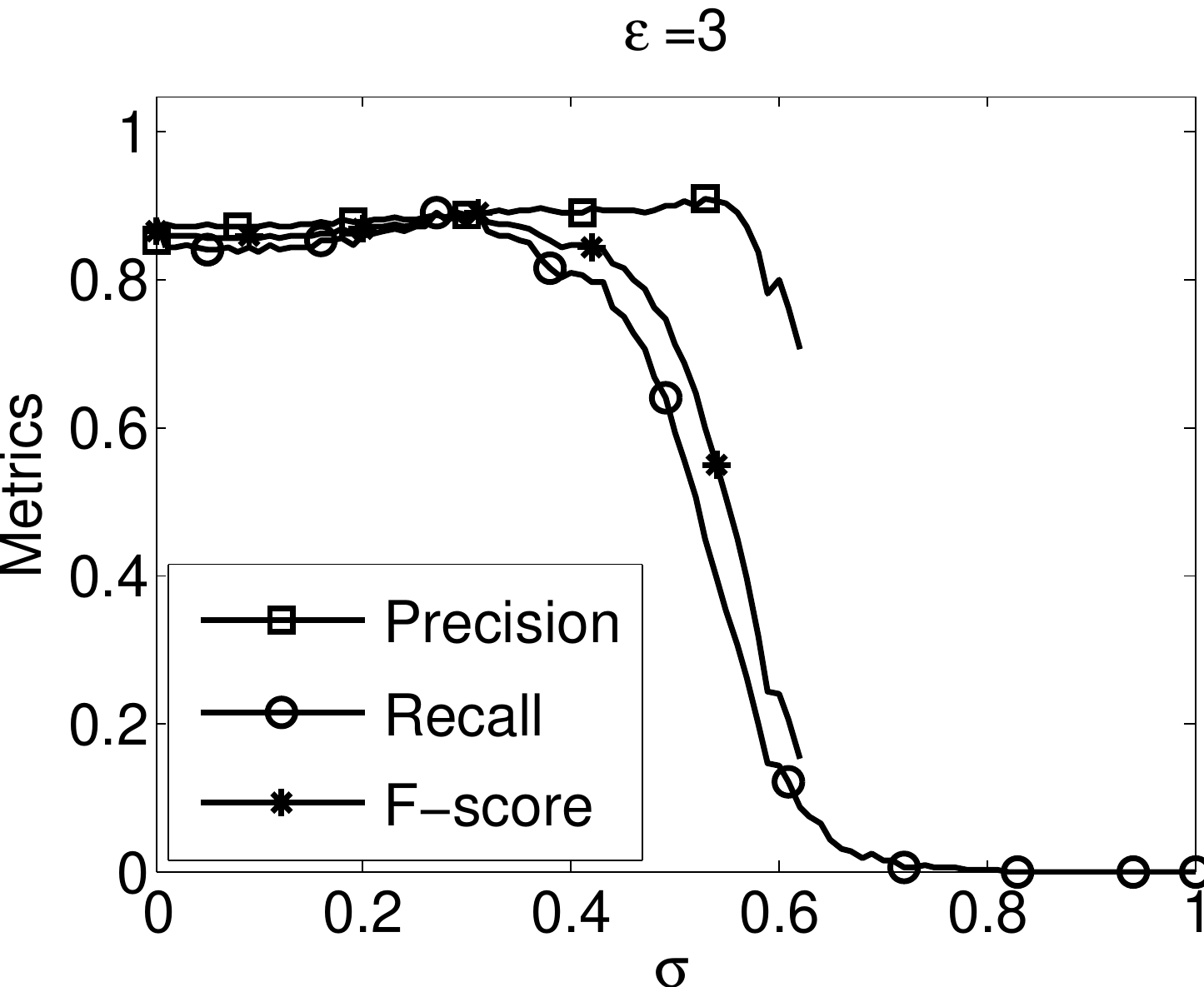}
        \label{fig:aggreg_art3:sl}
    }
    \end{subfloat}
    \begin{subfloat}[\textit{By Overlapping}]{
        \includegraphics[trim=1.7cm 7.cm 1.3cm 7.5cm, clip=true, width=100pt]{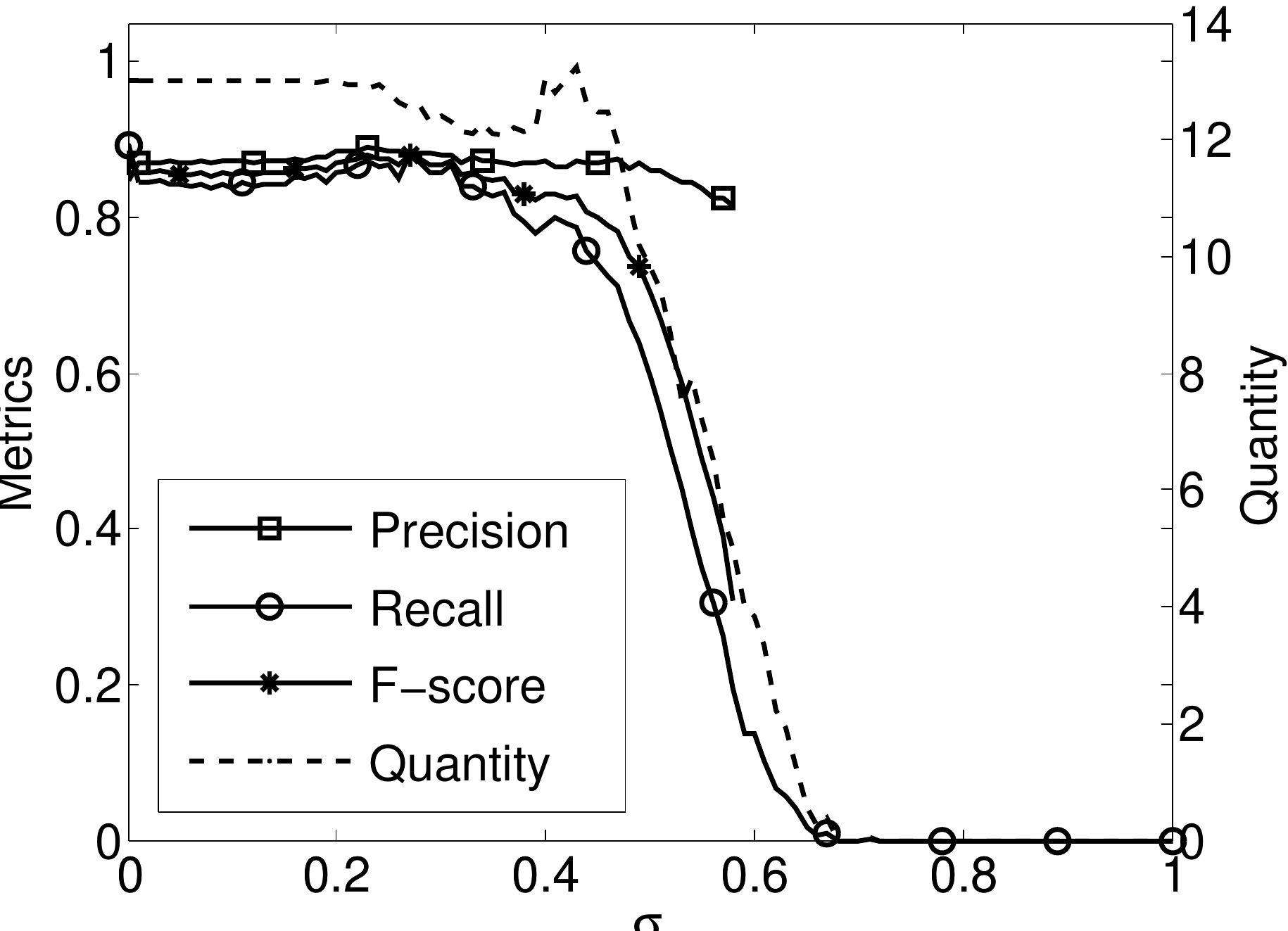}
        \label{fig:aggreg_art3:ov}
    }
    \end{subfloat}
    \begin{subfloat}[\textit{CE}]{
        \includegraphics[trim=2.9cm 7.7cm 2.7cm 7.5cm, clip=true, width=100pt]{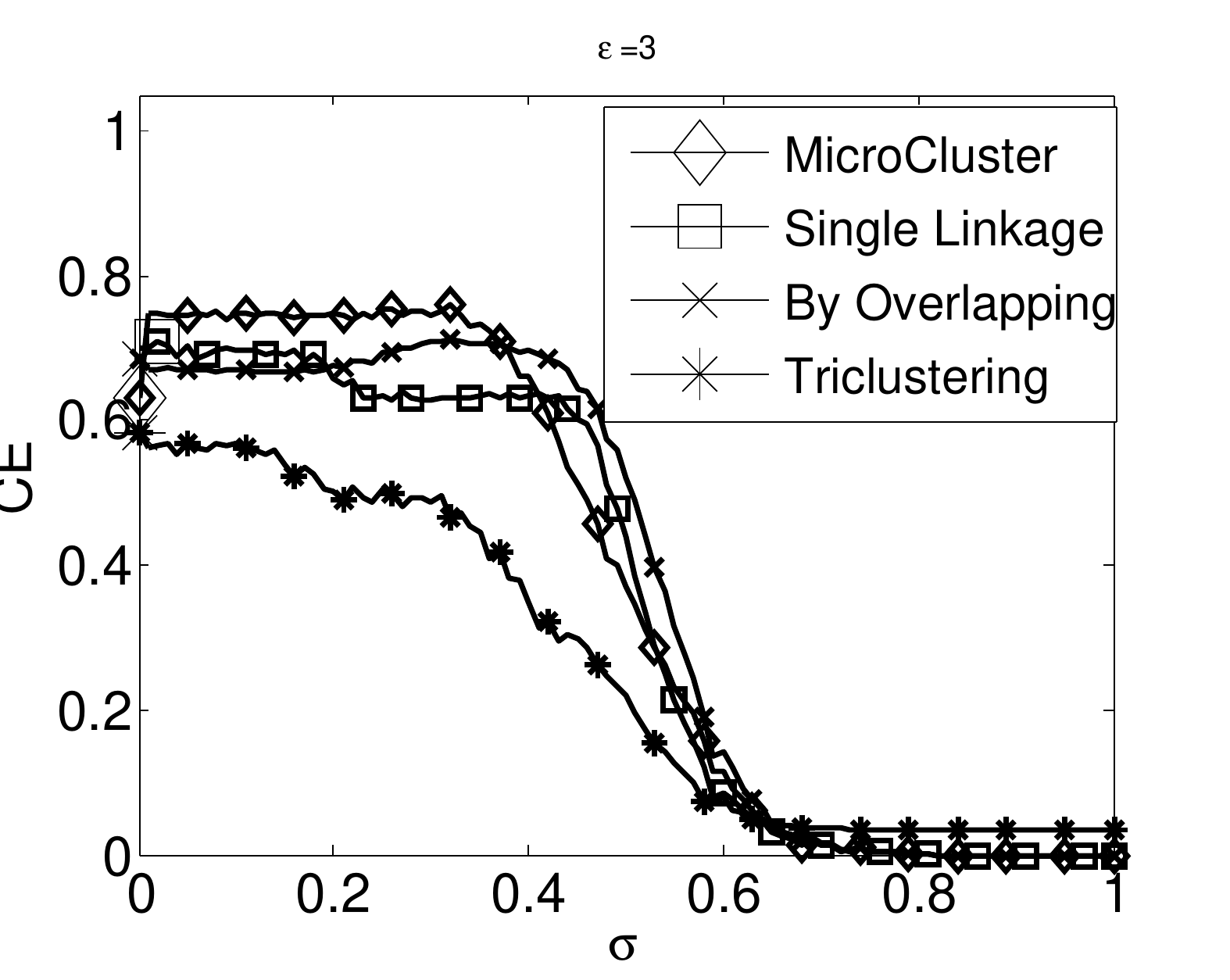}
        \label{fig:aggreg_art3:ces}
    }
    \end{subfloat}
    \caption{Solutions of aggregation as a function of the variance of the noise in dataset \textit{art3}.
    The scale on the right refers to quantity.}
    \label{fig:aggreg_art3}
\end{figure}
Figures \ref{fig:aggreg_art3:sl} and \ref{fig:aggreg_art3:ov} show the quality of aggregation with single linkage and by overlapping, respectively.
We can see that this dataset is more challenging than the previous ones.
However, the aggregation was able to significantly reduce the quantity of biclusters, while keeping a good quality.
Figure \ref{fig:aggreg_art3:ces} shows the CE metric for all aggregation methods.
Initially MicroCluster had a better performance, but our proposals were more robust to noise, getting a better result when $\sigma \gtrapprox 0.4$.

The aggregation was not only able to reduce the quantity of biclusters of the enumeration, but also improve the quality of the final result.
Now we are going to verify the behavior of the aggregation in real datasets.

\subsection{Experiments on Real Datasets}
We will start with the \textit{GDS2587} dataset by running RIn-Close to enumerate its coherent values biclusters.
We set $minRow = 50, minCol = 4$. When $\epsilon < 2.8$ no biclusters were found, and when $\epsilon = 3.0$ the quantity of biclusters was already huge.
We found 23, 2.825 and 19.649 biclusters when $\epsilon = 2.8, 2.9,$ and $3.0$, respectively.
\begin{figure}[]
    \centering
    \subfloat[$\epsilon = 2.8$]{\label{exp4:28_dendrogram}\includegraphics[trim=3cm 7.5cm 2.5cm 7cm, clip=true, width=100pt]{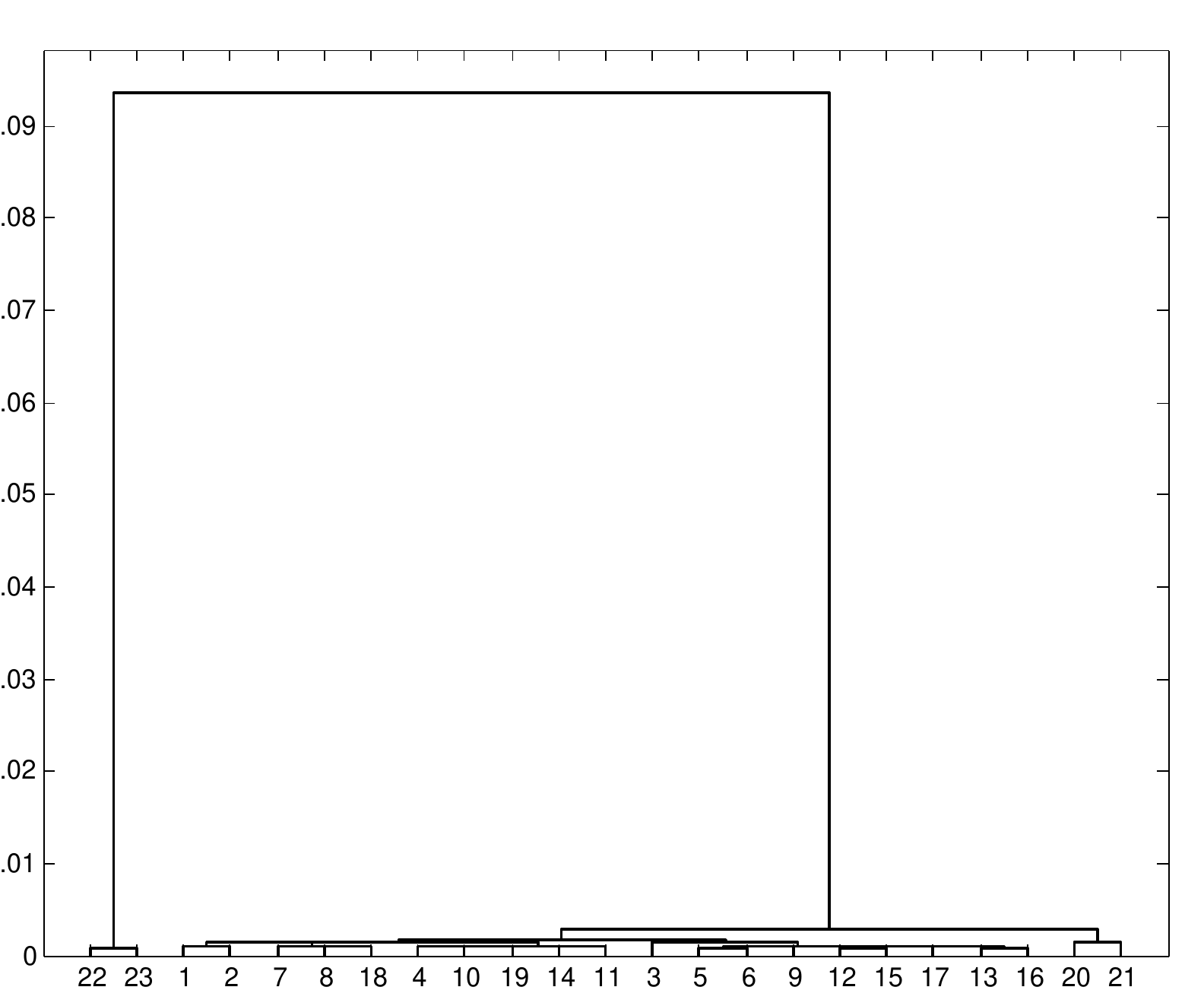}}
    \subfloat[$\epsilon = 2.9$]{\label{exp4:29_dendrogram}\includegraphics[trim=3cm 7.5cm 2.5cm 7cm, clip=true, width=100pt]{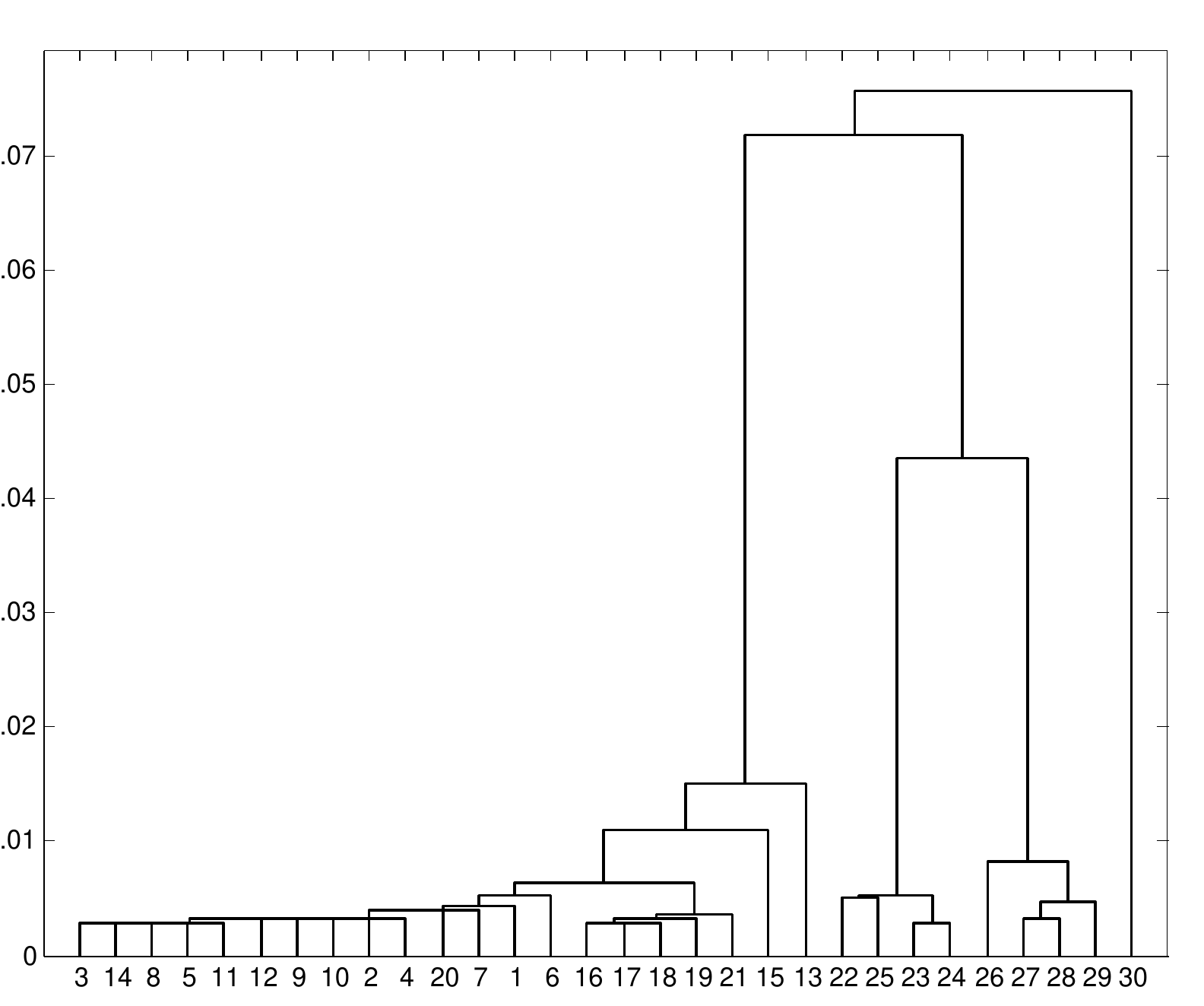}}
    \subfloat[$\epsilon = 3$]{\label{exp4:30_dendrogram}\includegraphics[trim=3cm 7.5cm 2.5cm 7cm, clip=true, width=100pt]{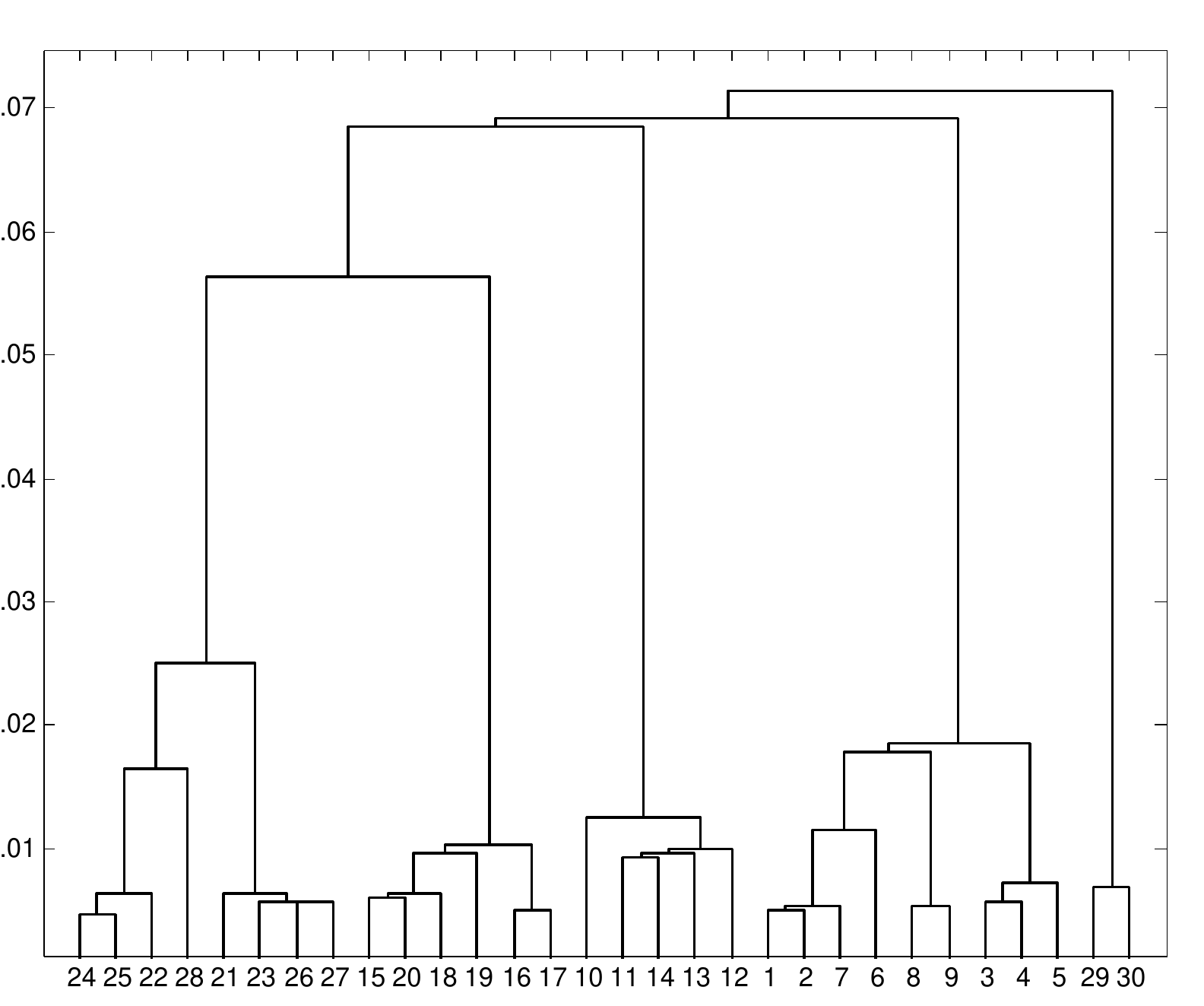}}
    \caption{Dendrograms of the aggregation with single linkage on \textit{GDS2587} dataset.}
    \label{fig:gds:dendrogram}
\end{figure}

Proceeding to the aggregation, Figure \ref{fig:gds:dendrogram} shows the dendrograms of the aggregation with single linkage.
In this case, the cuts are straightforward, having 2, 4, and 5 clusters respectively.
The aggregation by overlapping with a rate of $75\%$ reached the same quantity of biclusters.
We used these quantities to parameterize the triclustering algorithm.
The results of the aggregation with MicroCluster were very similar, and they depended only on the $\gamma$ parameter.
We got 7, 8 and 11 biclusters when $\gamma = 0.15, 0.1,$ and $0.05$, respectively.
We will now compare the results with the \textit{gene ontology enrichment analysis}.
A bicluster is called 'enriched' when any ontology term gets a p-value less than 0.01.

When $\epsilon = 2.8$, except for triclustering (only the first bicluster was enriched), all the algorithms returned only enriched biclusters.
In fact, the four main enriched terms were always the same, sometimes on different orders but with very close p-values.

\begin{table}[h]
\centering
\begin{tabular}{|l|l|l|p{7.3cm}|}
\cline{1-4}
GO Term     & p-val 		& counts 	& definition \\ \cline{1-4}
GO:0044464 & 0.00000000 & 39 / 774 & Any constituent part of a cell, the basic structural and functional unit of all organisms... \\ \cline{1-4}
GO:0044444 & 0.00000011 & 19 / 608 & Any constituent part of the cytoplasm, all of the contents of a cell excluding the plasma membrane...\\ \cline{1-4}
GO:0044424 & 0.00000350 & 19 / 578 & Any constituent part of the living contents of a cell; the matter contained within (but not including) the plasma membrane...\\ \cline{1-4}
\end{tabular}
\caption{Enrichment analysis of one bicluster from the aggregation by overlapping with rate of 70\%, on \textit{GDS2587} dataset.}
\label{tab:go}
\end{table}

When $\epsilon = 2.9$, all algorithms returned only enriched biclusters, including triclustering.
When $\epsilon = 3$, all algorithms except for triclustering returned only enriched biclusters.
Triclustering returned 4 from 5 enriched biclusters.

Table \ref{tab:go} shows the main enriched terms of one bicluster from the aggregation by overlapping after outlier removal, when $\epsilon = 2.8$.
In this case, the expert should choose which solution fits better the goal of the data analysis.

We will now proceed to the analysis of the \textit{FOOD} dataset.
We are going to verify how the aggregation changes the coverage of the dataset when compared to the enumeration.
As the aggregation will severely reduce the quantity of final biclusters, it is important to see if it will leave uncovered areas that were previously covered.

We replicated the experiment from Veroneze \textit{et al}. \cite{Veroneze2014} on this dataset and we will use $\epsilon = 1.25$ as recommended on that work.
With $minRow = 48, minCol = 2$ and looking for coherent values biclusters, the quantity of enumerated biclusters for $\epsilon = 1.25$ is 8.676.
\begin{figure}[]
    \centering
    \includegraphics[trim=2.8cm 8cm 2.9cm 7.5cm, clip=true, width=100pt]{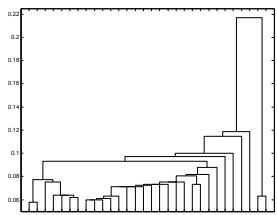}\vspace{11pt}
    \caption{Dendrogram for the aggregation with single linkage when $\epsilon = 1.25$ on \textit{FOOD} dataset.}
    \label{fig:food:dendrogram}
\end{figure}

Figure \ref{fig:food:dendrogram} shows the dendrogram of the aggregation with single linkage.
We can see that the cuts between 2 and 7 are acceptable.
In fact, cutting in two groups seems the best option, but it may be considered a small quantity of biclusters.
As from 4 to 5 the height is more pronounced, for the comparison it seems acceptable to cut the dendrogram on 4 objects.
The aggregation by overlapping with a rate of 70\% was also able to recover 4 aggregated biclusters.

MicroCluster with the deleting operation first was not able to properly aggregate the biclusters, keeping more than 800 biclusters when $\eta = 0.15$.
This behavior is the opposite of what happened with the artificial datasets.
There, when the deleting operation came first the results were more effective.
Here when the merging operation came first, the aggregation was able to reach 13 to 27 biclusters, depending on the $\gamma$ parameter.
As on the artificial datasets the best parameters were $\eta = \gamma = 0.15$, for the comparison we will use this parameterization with the merging operation occurring first, that gives us 13 biclusters.
For the triclustering algorithm we set $k = 4$, using insider information from the aggregation by overlapping.
\begin{table}[h]
\centering
\begin{tabular}{|l|l|l|l|l|}
\cline{1-5}
                  & Single Linkage & MicroCluster & Triclustering & RIn-Close\\ \cline{1-5}
By Ov.            & 12.50\% & 35.50\% & 70.31\% & 9.1\%\\ \cline{1-5}
Single Linkage    &  -      & 46.60\% & 81.51\% & 20.17\%\\ \cline{1-5}
MicroCluster      &  -      & -       & 45.73\% & 27.38\%\\ \cline{1-5}
Triclustering     &  -      & -       & -       & 61.33\%\\ \cline{1-5}
\end{tabular}
\caption{Difference in coverage of the solutions with the enumeration on \textit{FOOD} dataset.}
\label{tab:comparison}
\end{table}
Table \ref{tab:comparison} shows the comparison of difference in coverage (see Eq. \ref{eq:diff_cov}) between the aggregated solutions with the enumerated solution from RIn-Close.
We can see that the triclustering algorithm produces the most distinct solution when compared with the enumerated solution obtained with RIn-Close, exhibiting $\approx 61.33\%$ of difference in coverage.
The solutions from the aggregation by overlapping and with single linkage are relatively close to each other, as on the artificial datasets, showing a difference in coverage of $\approx 12.50\%$.
At the end, the closest solution to the RIn-Close results was the aggregation by overlapping, with a difference in coverage of $9.1\%$.
If we consider that this solution reduced the quantity of biclusters from 8.676 to 4 biclusters, the difference in coverage of only $9.1\%$ seems very promising.

\section{Considering Remarks and Future Work}
\label{conclusions}
We have compared the performance of our proposals against the most similar proposal in the literature, using artificial and real datasets.
The artificial datasets were characterized by a controlled structure of biclusters and were useful to show that the aggregation can severely reduce the quantity of biclusters, while increasing the quality of the final solution.
Our proposals outperformed the compared algorithms on the first two artificial datasets, and showed to be more robust to noise on the third artificial dataset.

We also verified if the aggregation could get enriched biclusters in the case of a gene expression dataset.
For different values of $\epsilon$ on the RIn-Close algorithm, we could see that the different methods of aggregation reached very similar results.
The main challenge of the aggregation with single linkage is to decide where to cut the dendrogram, but as we could see, this task was straightforward on the tested datasets.
Except for the triclustering, all aggregations returned only enriched biclusters.
And finally, we applied the aggregation methods to the \textit{FOOD} dataset and analyzed how the aggregation changed the coverage area when compared to the enumeration without aggregation.
Triclustering led to the most distinct result, and the aggregation by overlapping covered an area very similar to the area covered by the enumeration.

We can conclude that the aggregation is strongly recommended when enumerating all biclusters from a dataset.
The aggregation will not only significantly reduce the quantity of biclusters, but will also reduce the fragmentation and increase the quality of the final result.
A post-processing step for outlier removal brings additional robustness to the methodology.
As a further step of the research, we can adapt our proposals to work on an ensemble configuration.
We can also extend this work to deal with time series biclusters, which require contiguous attributes.

The authors would like to thank CAPES and CNPq for the financial support.
\bibliography{tese}

\begin{thebibliography}{10}

\bibitem{Cheng2000}
Y.~Cheng and G.~M. Church, ``{Biclustering of expression data.},'' {\em
  Proceedings of International Conference on Intelligent Systems for Molecular
  Biology; ISMB. International Conference on Intelligent Systems for Molecular
  Biology}, vol.~8, pp.~93--103, 2000.

\bibitem{Yang2003}
Y.~Jiong, H.~Wang, W.~Wang, and P.~Yu, ``Enhanced biclustering on expression
  data,'' in {\em Bioinformatics and Bioengineering, 2003. Proceedings. Third
  IEEE Symposium on}, pp.~321--327, 2003.

\bibitem{Makino2014}
K.~Makino and T.~Uno, ``New algorithms for enumerating all maximal cliques.,''
  in {\em SWAT} (T.~Hagerup and J.~Katajainen, eds.), vol.~3111 of {\em Lecture
  Notes in Computer Science}, pp.~260--272, Springer, 2004.

\bibitem{Uno2004}
T.~Uno, M.~Kiyomi, and H.~Arimura, ``Lcm ver. 2: Efficient mining algorithms
  for frequent/closed/maximal itemsets,'' in {\em FIMI}, vol.~126, 2004.

\bibitem{Andrews2009}
S.~A., ``In-close, a fast algorithm for computing formal concepts,'' in {\em
  the Seventeenth International Conference on Conceptual Structures}, 2009.

\bibitem{Veroneze2014}
R.~Veroneze, A.~Banerjee, and F.~J.~V. Zuben, ``Enumerating all maximal
  biclusters in real-valued datasets,'' {\em arXiv:1403.3562v3},
  vol.~abs/1403.3562, 2014.

\bibitem{Pandey2009}
G.~Pandey, G.~Atluri, M.~Steinbach, C.~L. Myers, and V.~Kumar, ``An association
  analysis approach to biclustering,'' in {\em Proceedings of the 15th ACM
  SIGKDD International Conference on Knowledge Discovery and Data Mining}, KDD
  '09, pp.~677--686, 2009.

\bibitem{Liu04}
J.~Liu, J.~Wang, and W.~Wang, ``Biclustering in gene expression data by
  tendency.,'' in {\em CSB}, pp.~182--193, IEEE Computer Society, 2004.

\bibitem{Zhao2005}
L.~Zhao and M.~J. Zaki, ``Microcluster: Efficient deterministic biclustering of
  microarray data,'' {\em {IEEE} Intelligent Systems}, vol.~20, no.~6,
  pp.~40--49, 2005.

\bibitem{Madeira2004}
S.~C. Madeira and A.~L. Oliveira, ``Biclustering algorithms for biological data
  analysis: A survey,'' {\em IEEE/ACM Trans. Comput. Biol. Bioinformatics},
  vol.~1, pp.~24--45, Jan. 2004.

\bibitem{Tanay2005}
T.~A., R.~Sharan, and R.~Shamir, ``Biclustering algorithms: A survey,'' in {\em
  In Handbook of Computational Molecular Biology Edited by: Chapman \& Hall/CRC
  Computer and Information Science Series}, 2005.

\bibitem{GAO2014}
T.~Gao and L.~Akoglu, ``Fast information-theoretic agglomerative
  co-clustering,'' in {\em Databases Theory and Applications} (H.~Wang and
  M.~A. Sharaf, eds.), vol.~8506 of {\em Lecture Notes in Computer Science},
  pp.~147--159, Springer International Publishing, 2014.

\bibitem{Hanczar2012}
B.~Hanczar and M.~Nadif, ``Ensemble methods for biclustering tasks,'' {\em
  Pattern Recognition}, vol.~45, no.~11, pp.~3938 -- 3949, 2012.

\bibitem{Hanczar2011b}
B.~Hanczar and M.~Nadif, ``{Improving the Biological Relevance of Biclustering
  for Microarray Data in Using Ensemble Methods},'' {\em 2011 22nd
  International Workshop on Database and Expert Systems Applications},
  pp.~413--417, Aug. 2011.

\bibitem{Horta2014}
D.~Horta and R.~J. G.~B. Campello, ``Similarity measures for comparing
  biclusterings,'' {\em Computational Biology and Bioinformatics, IEEE/ACM
  Transactions on}, vol.~11, pp.~942--954, Sept 2014.

\bibitem{Salton1971}
G.~Salton, ``Evaluation parameters,'' {\em The SMART Retrieval System,
  Experiments in Automatic Document Processing}, pp.~55--112, 1971.

\bibitem{vanRigsbergen1979}
C.~J. van Rigsbergen, {\em Information Retrieval}.
\newblock Englewood Cliffs: Prentice Hall, 1979.

\bibitem{Menestrina2009}
D.~Menestrina, S.~E. Whang, and H.~Garcia-Molina, ``Evaluating entity
  resolution results (extended version),'' technical report, Stanford
  University, 2009.

\bibitem{Prelic2006}
A.~Preli\'{c}, S.~Bleuler, P.~Zimmermann, A.~Wille, P.~B\"{u}hlmann,
  W.~Gruissem, L.~Hennig, L.~Thiele, and E.~Zitzler, ``A systematic comparison
  and evaluation of biclustering methods for gene expression data,'' {\em
  Bioinformatics}, vol.~22, pp.~1122--1129, May 2006.

\end{thebibliography}
\end{document}